\documentclass[sigconf,natbib=false]{acmart}
\usepackage{multirow}
\usepackage{booktabs}
\usepackage{colortbl}
\usepackage[linesnumbered,ruled,vlined]{algorithm2e}

\AtBeginDocument{%
  }

\copyrightyear{2024}
\acmYear{2024}
\setcopyright{none}
\acmConference[MM '24]{Proceedings of the 32nd ACM International Conference on Multimedia}{October 28-November 1, 2024}{Melbourne, VIC, Australia}
\acmBooktitle{Proceedings of the 32nd ACM International Conference on Multimedia (MM '24), October 28-November 1, 2024, Melbourne, VIC, Australia}
\acmDOI{xxxxx}
\acmISBN{xxxxx}

\RequirePackage[
  datamodel=acmdatamodel,
  style=acmnumeric,
  ]{biblatex}
\begin{document}

%%
%% The "title" command has an optional parameter,
%% allowing the author to define a "short title" to be used in page headers.
\title{Sample-agnostic Adversarial Perturbation for Vision-Language Pre-training Models}

%%
%% The "author" command and its associated commands are used to define
%% the authors and their affiliations.
%% Of note is the shared affiliation of the first two authors, and the
%% "authornote" and "authornotemark" commands
%% used to denote shared contribution to the research.
\author{Haonan Zheng}
\email{zhenghaonan@mail.nwpu.edu.cn}
\orcid{0009-0009-2017-962X}
\affiliation{%
  \institution{Northwestern Polytechnical University}
  \department{School of Electronics and Information}
  \city{Xi'an}
  \state{Shaanxi}
  \country{China}
}

\author{Wen Jiang}
% \authornotemark[2]
% \email{jiangwen@nwpu.edu.cn}
% \orcid{0000-0001-5429-2748}
\affiliation{%
  \institution{Northwestern Polytechnical University}
  \department{School of Electronics and Information}
  \city{Xi'an}
  \state{Shaanxi}
  \country{China}
}

\author{Xinyang Deng}
\authornote{Corresponding author}
% \email{xinyang.deng@nwpu.edu.cn}
% \orcid{0000-0001-8181-7001}
\affiliation{%
  \institution{Northwestern Polytechnical University}
  \department{School of Electronics and Information}
  \city{Xi'an}
  \state{Shaanxi}
  \country{China}
}

\author{Wenrui Li}
\email{liwr618@163.com}
\orcid{0000-0002-2393-9016}
\affiliation{%
  \institution{Harbin Institute of Technology}
  \department{Department of Computer Science and Technology}
  \city{Harbin}
  % \state{Shaanxi}
  \country{China}
}

%%
%% By default, the full list of authors will be used in the page
%% headers. Often, this list is too long, and will overlap
%% other information printed in the page headers. This command allows
%% the author to define a more concise list
%% of authors' names for this purpose.
\renewcommand{\shortauthors}{Haonan Zheng, Wen Jiang, Xinyang Deng, and Wenrui Li}

%%
%% The abstract is a short summary of the work to be presented in the
%% article.
\begin{abstract}
% Recently, Vision-Language Pre-training (VLP) models have shown vulnerable to imperceptible, well sought, perturbations of images or texts. 
% However, the perturbations obtained by existing methods are sample-specific, that is a separate perturbation needs to be generated for each sample.
% In this paper, we present the first study to construct sample-agnostic perturbation based on decision boundaries. 
% We first investigate how to push sample point across the decision boundaries of linear multi-class classifier under the top-k accuracy metric.
% Based on this theoretical foundation, for visual-language tasks, we treat the visual and textual modalities as mutually serving as sample points and decision boundaries.
% We devise algorithms to guide an image embedding to cross the decision boundaries constructed by text embeddings, and vice versa.
% Throughout the process, we consistently optimize a single universal perturbation, ultimately identifying a singular direction in the input space that can be exploited to compromise the VLP model.
% The algorithms proposed support the generation of universal perturbation in the form of global noise or adversarial patch.
% Our experiments demonstrate the data transferability, task transferability, and model transferability of our attack method across various VLP models and datasets.
Recent studies on AI security have highlighted the vulnerability of Vision-Language Pre-training (VLP) models to subtle yet intentionally designed perturbations in images and texts. Investigating multimodal systems' robustness via adversarial attacks is crucial in this field. Most multimodal attacks are sample-specific, generating a unique perturbation for each sample to construct adversarial samples. To the best of our knowledge, it is the first work through multimodal decision boundaries to explore the creation of a universal, sample-agnostic perturbation that applies to any image. Initially, we explore strategies to move sample points beyond the decision boundaries of linear classifiers, refining the algorithm to ensure successful attacks under the top $k$ accuracy metric. Based on this foundation, in visual-language tasks, we treat visual and textual modalities as reciprocal sample points and decision hyperplanes, guiding image embeddings to traverse text-constructed decision boundaries, and vice versa. This iterative process consistently refines a universal perturbation, ultimately identifying a singular direction within the input space which is exploitable to impair the retrieval performance of VLP models. The proposed algorithms support the creation of global perturbations or adversarial patches. Comprehensive experiments validate the effectiveness of our method, showcasing its data, task, and model transferability across various VLP models and datasets.
Code: https://github.com/LibertazZ/MUAP
% In VLP models, our primary focus is on the vulnerability of contrastive representations (two unimodal [cls] embeddings), and fused representation (a multimodal [cls] embedding).
% The initial intention of the former is to learn image representations directly from raw text, based on the Image-Text Contrast (ITC) task.
% Following this perspective, we treat text as the label, using text representations to partition the feature space, thereby reducing ITC to an image classification problem. 
% The latter is commonly employed in executing Image-Text Matching (ITM) task. Viewing unimodal encoders as modeling marginal probability and fusion encoder as modeling joint probabilities, the ITM task can be seen as modeling a conditional probability image classification problem with text as the conditional variable. 
% We investigate the efficacy of various adversarial attack methods across multiple VLP models and datasets in diverse V+L downstream tasks. This establishes a new baseline for white-box robustness studies in multimodal scenario.

\end{abstract}

%%
%% The code below is generated by the tool at http://dl.acm.org/ccs.cfm.
%% Please copy and paste the code instead of the example below.
%%


%% Keywords. The author(s) should pick words that accurately describe
%% the work being presented. Separate the keywords with commas.

% \received{20 February 2007}
% \received[revised]{12 March 2009}
% \received[accepted]{5 June 2009}

%%
%% This command processes the author and affiliation and title
%% information and builds the first part of the formatted document.
\maketitle

\section{Introduction}

Multimodal contrastive learning \cite{clip, mrl} aims to leverage raw text as a much broader source of supervision to encourage models to learn more flexible visual representations which align with textual representations.
% √
Building on this, by incorporating masked data modelling \cite{bert, beit,beit2,beit3}, image-text matching \cite{vilt}, and other advanced pre-training tasks \cite{tcl}, a multimodal learning paradigm based on pre-training plus fine-tuning has gradually formed.
% √
This paradigm shift has facilitated the inception of various VLP models \cite{clip,blip,albef,vlmo,li1} featuring meticulously crafted architectures and adaptable pre-trained weights which can be fine-tuned \cite{cocoop, coop} to perform a wide range of downstream tasks \cite{vqa2,itr1, itr2, itr3}.
% √
Behind these promising outcomes lie potential security risks \cite{coattack,fgsm,pgd,rq1}.
% √
Research indicates that adding imperceptible noise to images or making minor adjustments at the word or letter level in texts can lead to severe performance degradation in VLP models \cite{sga, advclip, vlbackdoor, zheng1}.
% √
These maliciously crafted inputs are adversarial examples \cite{fgsm}.
% √
Existing work has explored the generation of adversarial examples in VLP models from multiple perspectives.
% √
For example, data poisoning can disrupt the training phase, resulting in models that are susceptible to backdoor attacks \cite{vlbackdoor}. 
% √
During testing, attacks are conducted either in the white-box scenario where the model weights are transparent \cite{coattack}, or in the black-box context where the model is inaccessible causing transferability to be the primary consideration \cite{sga}.
% √
% This work primarily focuses on the testing phase, which means we do not consider poisoned models.
Given the public availability of pre-trained weights, we aim to design a novel white-box attack method for pre-trained VLP models like CLIP \cite{clip}, exploring potential security vulnerabilities in the multimodal domain.
% Given the public availability of the pre-trained weights, we are dedicated to designing a novel white-box attack method on pre-trained VLP models, such as CLIP \cite{clip}, to explore potential security vulnerabilities in the multimodal domain.
% √
% Diverging from most existing attack strategies, our approach incorporates decision boundary theory to craft a sample-agnostic perturbation which can be directly applied to any sample to generate adversarial examples.
% % √
We further consider the attack patterns from feasibility, immediacy and practicality.
% √
(1) \textbf{Feasibility}: 
% Due to the differentiability (continuous pixel values) and semantic robustness (minor perturbations don't affect human perception), image inputs are suitable carriers for malicious noise, whereas attacking textual modality usually involves operations such as word replacement, which is inevitable to cause semantic errors.
The differentiability (continuous pixel values) and semantic robustness (minor perturbations don’t impact human perception) make image inputs apt carriers for malicious noise. 
In contrast, attacking text often involves operations like word replacement, inevitably leading to semantic errors.
% √
In application scenarios, text often mediates human-computer interactions, and attackers have no access to texts directly provided by users.
% √
Conversely, images often automatically extracted by intelligent models are accessible for attackers to introduce perturbations. 
% √
% Furthermore, we highlight the perturbation's universality, which, while challenging to optimize for non-differentiable textual tokens, is relatively feasible for image inputs.
Furthermore, we emphasize the universality of perturbations. 
Although optimizing them for non-differentiable textual tokens is challenging, it is relatively feasible for image inputs.
% √
Therefore, we focus on attacking the visual modality, without neglecting multimodal interactions.
% √
(2) \textbf{immediacy}:
The expeditious process of image acquisition by visual sensors contrasts with sample-specific attacks requiring the generation of adversarial perturbations for images on the spot.
% √
Diverging from sample-specific strategies, our approach incorporates decision boundary theory to craft a sample-agnostic perturbation.
% √
Irrespective of various images captured at any moment, directly applying a prepared perturbation can create effective adversarial images.
% √
Therefore, the primary consideration is universality.
% √
% which can be directly applied to any sample to generate adversarial examples.
% √
% The expeditious process of image acquisition by visual sensors contrasts with sample-specific attacks requiring to generate adversarial perturbations for images on the spot, whereas sample-agnostic attacks can pre-emptively identify a universal adversarial direction in the input space.
(3) \textbf{Practicality}: 
Image perturbation can be categorized into two main types: global noise and adversarial patch. 
% √
% Global noise is impractical for real-world applications, as directly modifying pixel values in physical scenarios is not feasible.
Global noise is impractical for real-world applications due to the infeasibility of directly modifying pixel values in physical scenarios.
% √
% In contrast, adversarial patches, which cover only a portion of an image, can be readily deployed in the physical world.
In contrast, adversarial patches cover only part of an image and can be easily deployed in the physical world.
% √
Therefore, our primary focus is patch form, with additional discussion on global perturbation for completeness.
% √

%%%%%%%%%%%%%%%%%%%%%%%%%%%%%%%%%%%%%%%%%%%%%%%%%%%%%%%%%%%%%%%%%%%%%%%%%
%  开始 Introduction图
%%%%%%%%%%%%%%%%%%%%%%%%%%%%%%%%%%%%%%%%%%%%%%%%%%%%%%%%%%%%%%%%%%%%%%%%%
\begin{figure}[t]
\centering
\includegraphics[width=\linewidth]{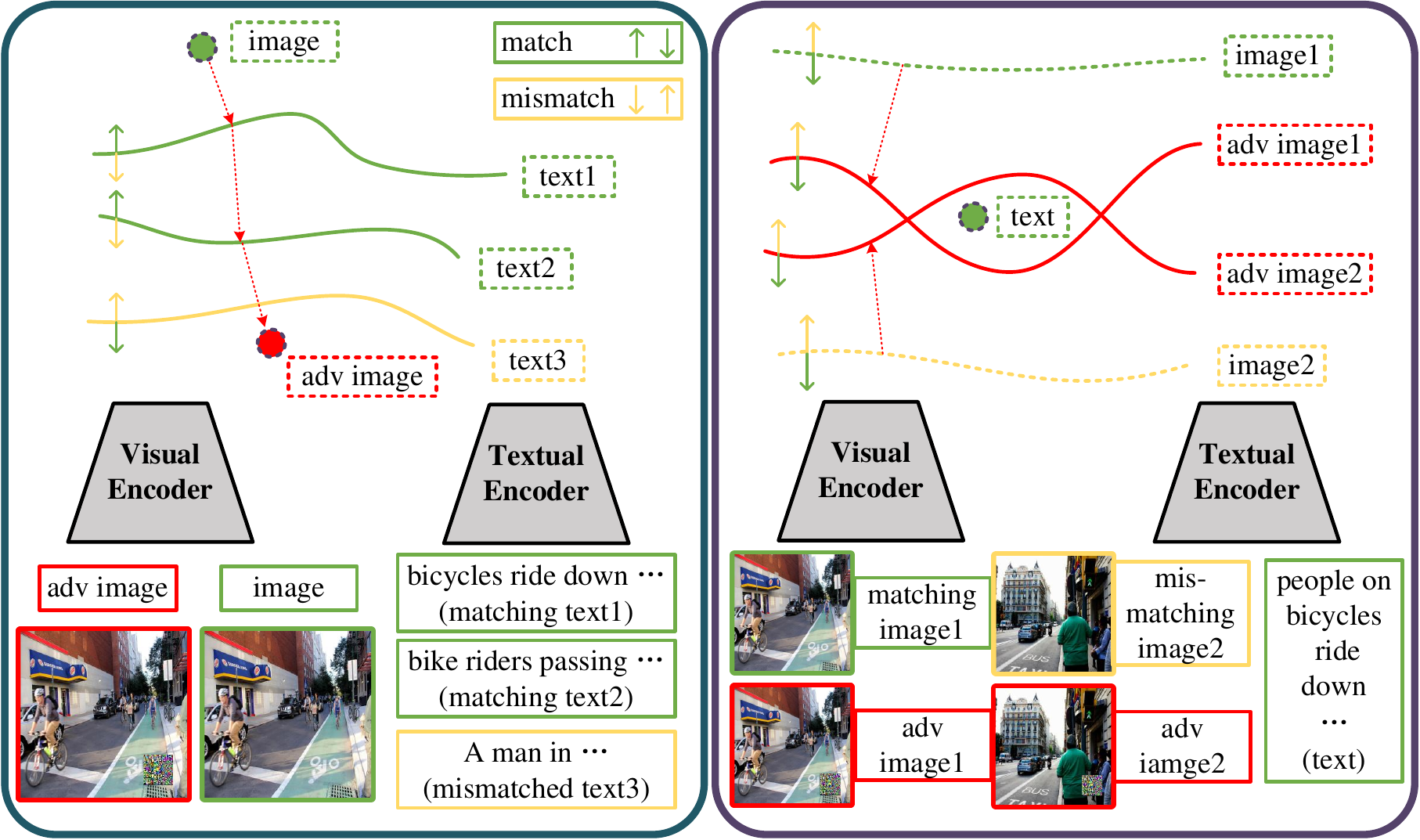}
\caption{
The left part represents attacking text retrieval, involving guiding benign embedding across decision boundaries to generate adversarial images; 
the right part represents attacking image retrieval, involving distorting benign decision hyperplanes to construct malicious boundaries, corrupting the decision outcomes for benign text.}
\label{fig:introduction}
\Description{The top part represents attacking image-to-text retrieval, involving guiding benign embedding across decision boundaries to generate adversarial images; 
the bottom part represents attacking text-to-image retrieval, involving distorting benign decision hyperplanes to construct malicious decision boundaries, corrupting the decision outcomes for benign text.}
\end{figure}

As mentioned above, we focus on designing the universal adversarial patch for image modality in the white-box scenario. 
% √
% To this end, we innovatively propose a generation algorithm of universal patches rooted in decision boundary theory, marking a novel approach in the multimodal domain.
To achieve this, we propose an innovative algorithm for generating universal patches based on decision boundary theory, introducing a novel approach in the multimodal domain.
% √
% Beginning with the foundational principles of the binary linear classifier, we delineate the calculations of the distance and direction of a point to the decision hyperplane. 
Starting with the basic principles of the binary linear classifier, we outline how to calculate the distance and direction from a point to the decision hyperplane.
% √
% This principle is expanded to accommodate the linear multi-class classifier, wherein we detail a strategy to compel sample points across multiple decision boundaries simultaneously, guaranteeing successful attacks under the top \(k\) accuracy metric.
We extend this principle to the linear multi-class classifier, detailing a strategy that forces sample points across multiple decision boundaries simultaneously, ensuring successful attacks under the top \(k\) accuracy metric.
% √
Subsequently, we apply these theoretical results to the multimodal scenario, in which images and texts can serve interchangeably as sample points and decision boundaries. 
% √
This gives rise to two attack methodologies as shown in Fig~\ref{fig:introduction}. 
% √
The key contributions of our research are as follows:
% √
\begin{itemize}
    \item 
    We pioneer the application of decision boundary theory to the multimodal scenario, providing a theoretical foundation for exploring the robustness of image and visual embeddings within the high-dimensional space.
    \item 
    We design a novel algorithm for generating universal adversarial perturbation, a singular malicious direction in the image input space that can disrupt retrieval performance.
    \item 
    Through comprehensive experimentation across various datasets and VLP models, our universal perturbation exhibits remarkable adaptability across samples, highlighting its efficacy in transfer from pre-trained encoders to downstream applications.
\end{itemize}

\section{Related Work}
% This section discusses related works about VLP models, decision boundaries and universal adversarial attacks.

\textbf{Vision-Language Pre-training Models.}
Initially, VLP models are developed leveraging pre-trained object detectors, employing region features to acquire vision-language representations \cite{uniter, oscar, vinvl, vlmixer, lwr3}.
% √
The Vision Transformer (ViT) \cite{vit} introduces a powerful transformer-based encoder for visual data, enabling unified feature extraction from multimodal inputs, coupled with multimodal contrastive learning, establishing a straightforward and efficient paradigm for multimodal pre-training.
% √
A representative model is CLIP \cite{clip}, which includes a text and visual encoder. 
% √
These unimodal encoders extract textual and visual representations and align them within a normalized contrastive space.
% √
Subsequently, additional pre-training tasks are introduced into the multimodal domain, such as Masked Language Modeling (MLM) proposed by BERT \cite{bert}, and Masked Image Modeling (MIM), proposed by BEiT \cite{beit} and further developed by BEiTV2 \cite{beit2}. 
% √
BEiT3 unifies these results into Masked Data Modeling (MDM), which integrates MIM, MLM, and Masked Language-Vision Modeling (MLVM), marking a new paradigm for multimodal pre-training.
% √
Based on these, incorporating Multiway Transformer \cite{vlmo} which splits the
feed-forward layer into three parallel paths for unimodal feature extraction and feature fusion respectively, BEiT3 achieves better performance across a range of downstream applications.
% √

\textbf{Decision Boundary.}
DeepFool \cite{deepfool} is an adversarial algorithm tailored for neural network image classifiers, leveraging decision boundaries to induce misclassification through minimal perturbations. 
% √
Although it offers an effective strategy based on decision boundary analysis, its application is limited to image classification tasks.
% √
The algorithm aims to achieve minimal perturbation by iteratively exploring the direction of proximal disturbance. 
% √
This process requires extensive gradient backpropagation, significantly increasing computational time, especially as the number of categories expands, making the approach less feasible.
% This process demands extensive gradient backpropagation, significantly increasing computational time, especially with category numbers expanding, which renders the approach less feasible.
% √
We adopt the decision boundary concept, extend it to the multimodal scenario, and introduce pragmatic solutions, e.g., utilizing cosine similarity to ascertain the nearest decision boundary, effectively reducing redundant gradient backpropagation to enhance efficiency.
% √

\textbf{Universal Adversarial Perturbation.}
Universal adversarial perturbations in image classification are extensively discussed and can be divided into sample-dependent and sample-independent.
% √
The former kind includes methods such as UAP\cite{uap}, SGUAP\cite{sguap}, NAG\cite{sguap}, GAP\cite{gap}, and FFF\cite{fff}, which access a set of surrogate images to identify a universal adversarial perturbation, utilizing generative adversarial networks or various optimization strategies.
% √
It is noteworthy that these methods focus exclusively on global perturbation. 
% √
Similarly, our approach requires a batch of proxy data, yet we prioritize the adversarial patch which is more practical in the physical world.
% √
The other kind involves methods that do not interact with training images, such as Data-free UAP\cite{dfuap}. 
% √
This approach generates class impressions for each category based on which it constructs universal perturbations. 
% √
Since class impressions are tailored specifically for the classification task, they are not applicable in the multimodal context.
% √
The work most relevant to ours is AdvCLIP\cite{advclip}, which has developed a universal patch attack for multimodal contrastive learning based on topological structures.
% √
It provides valuable insights into addressing the modality gap, boosting attack transferability between pre-trained encoders and downstream tasks, as well as some defence plans.
% √
However, it relies on Generative Adversarial Networks (GANs), incurring additional overhead, and the topological structure loss fails to generate stable attack effects. 
% √
Furthermore, it lacks validation on large-scale datasets and models.
% √

\section{Methodology}

\subsection{Theoretical Basis}
% This section uses simple linear classifiers as an example to explain the theoretical basis of our method.
% √

\begin{figure}[t]
\centering
\includegraphics[width=\linewidth]{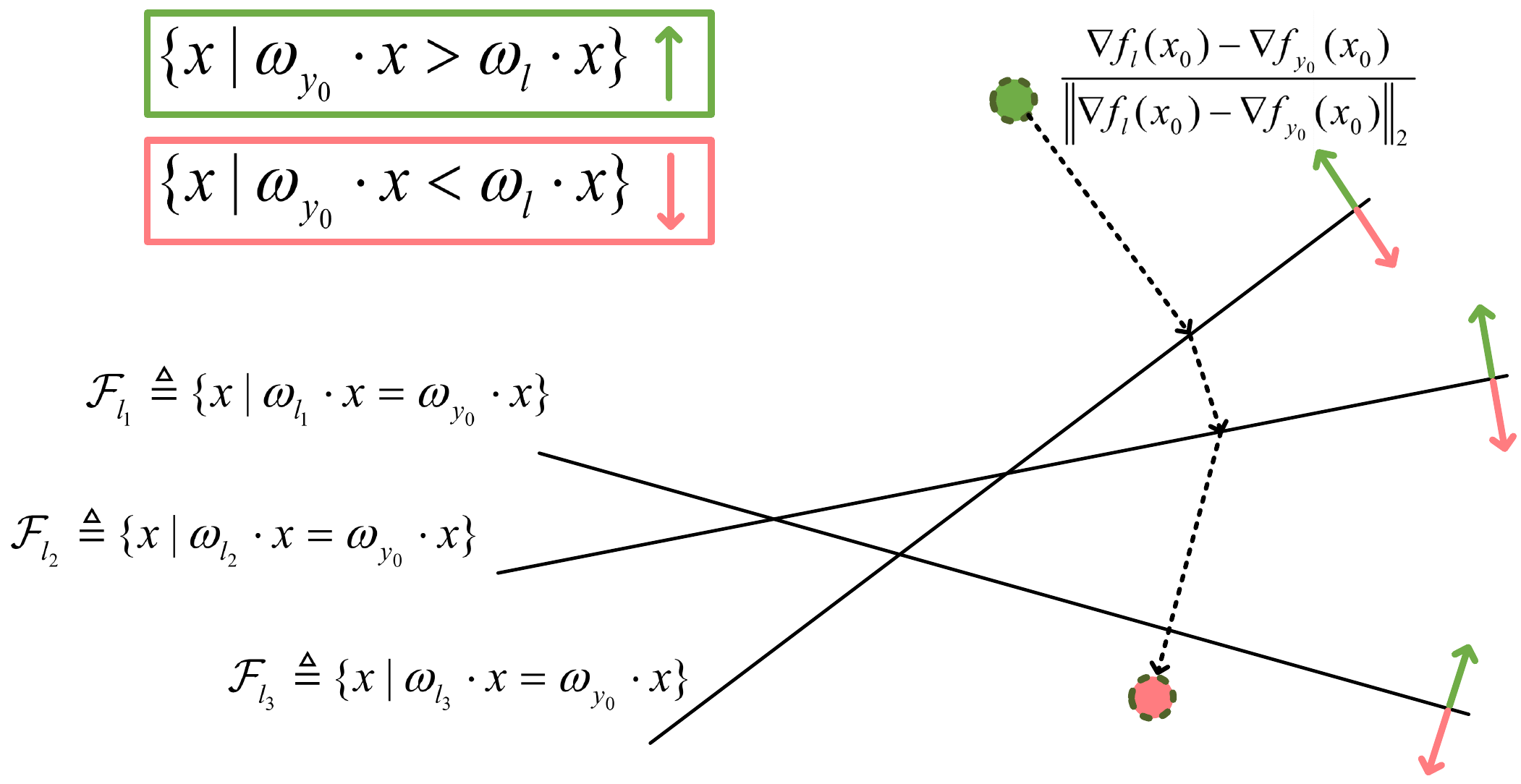}
\caption{Illustration of the disturbance trajectory that simultaneously crosses the top \(k\) decision boundaries.}
\label{fig:method}
\Description{Illustration of perturbation direction and perturbation distance for a binary linear classifier.}
\end{figure}

\textbf{Linear Binary Classifier.} 
We consider a sample pair as \((x, y)\), where \(x \in \mathbb{R}^n\) is the input to be classified, and \(y \in \{-1, +1\}\) is the true label. 
A binary classifier is given by \(\hat{y} = \text{sign}(f(x))\), where \(f: \mathbb{R}^n \rightarrow \mathbb{R}\) is a linear function \(f(x) = \omega \cdot x + b\), and \(\hat{y} \in \{-1, +1\}\) is the predicted category.
This binary classifier divides the n-dimensional space into two parts: \(\{x | f(x) > 0\}\) and \(\{x | f(x) < 0\}\), with the decision boundary being the hyperplane \(\mathcal{F} \triangleq \{x | w \cdot x + b = 0\}\).
Considering a sample \(x_0\) where \(f(x_0) > 0\), i.e. \(\hat{y}_0 = +1\), we try to find a perturbation \(r\) which will be added to the clean sample \(x_0\) to obtain the adversarial sample \(x_0^{adv} = x_0 + r\). 
According to the point-to-plane distance formulation, the distance from \(x_0\) to the decision boundary \(\mathcal{F}\) is: 
\begin{equation}
\frac{f(x_0)}{\left \| \omega \right \|_2 } = \frac{f(x_0)}{\left \| \nabla f(x_0) \right \|_2 }
\end{equation}
where \(\nabla{f(x_0)}\) represents the gradient of the function \(f\) with respect to \(x_0\).
The direction of perturbation \(r\) is normal to the decision boundary \(\mathcal{F}\), which is:
\begin{equation}
-\frac{\omega}{\left \|\omega\right \|_2} =-\frac{\nabla f(x_0)}{\left \|\nabla f(x_0)\right \|_2}
\end{equation}
Based on this, the perturbation direction \( r \) is given by distance times unit normal vector:
\begin{equation}
r = -\frac{f(x_0) \cdot \nabla f(x_0) }{\|\nabla f(x_0)\|_2^2}
\end{equation}
where \(\|\cdot\|_2\) denotes the Euclidean distance.

\textbf{Linear multi-class classifier.} 
We continue to consider a multi-class classifier (such as a linear layer), where \((x, y)\) is a sample pair, with \(y \in \{1, 2, \ldots, C\}\).
% Let \(E: \mathbb{R}^n \rightarrow \mathbb{R}^d\) be an encoder that projects the input \(x\) into a \(d\)-dimensional decision space.
The weight matrix of the linear layer is \(W = \{\omega_i\}_{i=1}^C\), where \(\omega_i \in \mathbb{R}^n\) is the feature vector corresponding to category \(i\).
We define \(f_i(x) =  \omega_i \cdot x \), where \(\cdot\) represents vector inner product. Then, the prediction result of the multi-class classifier is \(\hat{y} = \arg\max_i(f_i(x))\).
We consider a correctly classified sample pair \((x_0, y_0)\), which satisfies:
\begin{equation}
\forall i \in \{1, 2, \ldots, C\}\ \wedge \ i \neq y_0, \, f_{y_0}(x_0) > f_i(x_0)
\end{equation}
To attack the multi-class classifier, we similarly need to find an adversarial perturbation \(r\) to push the sample \(x_0\) across the decision boundary. 
In this scenario, there exist \(C-1\) decision boundaries \(\mathcal{F}_i \triangleq \{x | f_{y_0}(x) - f_i(x) = 0\}=\{x|(\omega_{y_0}-\omega_i)\cdot x =0\}\) (\(i \neq y_0\)). 
\(x_0\) crossing any decision boundary will lead to misclassification. 
Therefore, we first select the nearest decision boundary \(\mathcal{F}_l\):
\begin{equation} \label{equ:l}
l=\underset{i \neq y_0}{argmin} \frac{f_{y_0}(x_0)-f_i(x_0)}{\left \| \omega_{y_0}-\omega_i \right \|_2 }= \underset{i \neq y_0}{argmin} \frac{f_{y_0}(x_0)-f_i(x_0)}{\left \| \nabla f_{y_0}(x_0)-\nabla f_i(x_0) \right \|_2}
\end{equation}
Based on this, we can determine that the nearest perturbation direction is normal to the decision boundary \(\mathcal{F}_l\), that is:
\begin{equation}
\begin{aligned}
-\frac{\omega_{y_0}-\omega_l}{\left \| \omega_{y_0}-\omega_l \right \|_2 }= \frac{\nabla f_l(x_0)-\nabla f_{y_0}(x_0) }{\left \| \nabla f_{y_0}(x_0)-\nabla f_l(x_0) \right \|_2}
\end{aligned}
\end{equation}
Finally, we can obtain the perturbation \(r\):
\begin{equation}
r = \frac{\left(\nabla f_l(x_0)-\nabla f_{y_0}(x_0)\right)\cdot  \left(f_{y_0}(x_0)-f_l(x_0)\right) }{ \left \| \nabla f_{y_0}(x_0)-\nabla f_l(x_0) \right \|_2^2}
\end{equation}

\textbf{Multi-class classifier under top \(k\) metric.} 
The described attack strategy for the multi-class classifier faces a problem: the adversarial sample \(\hat{x}_0\) crosses only the nearest decision boundary.
If the evaluation is based on the top-\(k\) metric (i.e., the classification result is considered correct if the true category is among the top \(k\) candidate categories with the highest scores), this attack scheme cannot guarantee a successful attack.
Therefore, the perturbation \(r\) must guide \(x_0\) across the \(k\) nearest decision boundaries simultaneously, ensuring the top \(k\) highest-scoring categories exclude the ground truth label.
% Therefore, we require the perturbation \(r\) to guide \(x_0\) across the \(k\) nearest decision boundaries simultaneously so that we can ensure that the top-\(k\) highest-scoring categories do not contain the ground truth label.
Based on Equation~\ref{equ:l}, we can determine the \(k\) nearest decision boundaries \(\{\mathcal{F}_{l_j}\}_{j=1}^{k}\). 
We obtain the perturbation \(r\) in an iterative process, as shown in Alg~\ref{alg:topkdeepfool}. 
We set \(\eta = 0.02\) representing the extent to cross the decision boundary in each iteration.

\IncMargin{1em}
\begin{algorithm}[t]
\SetKwData{Left}{left}\SetKwData{This}{this}\SetKwData{Up}{up} \SetKwFunction{Union}{Union}\SetKwInOut{Input}{Input}\SetKwInOut{Output}{Output}	
\Input{Clean simple pair $(x,y)$, \(k\) nearest decision boundaries \(\{\mathcal{F}_{l_j}\}_{j=1}^{k}\), \(L=\{l_j\}_{j=1}^{k}\)} 
\Output{Perturbation \(r\).} 
Initialize \(x_0 \gets x\), \(i \gets 0\), \(r_0 \gets \boldsymbol {0}\)
\BlankLine 
% \emph{special treatment of the first line}\; 
\While{\(\exists l\in L, \, f_{y}(x+(1 + \eta) \cdot r)> f_{l}(x+(1 + \eta) \cdot r)\)}
{
\(\hat{x} \gets x + r\)

\(l_{min} \gets \underset{l\in L}{argmin} \frac{f_{l}(\hat{x})-f_y(\hat{x})}{\left \| \nabla f_{l}(\hat{x})-\nabla f_y(\hat{x}) \right \|_2}\)

\(r \gets r + \frac{\left(\nabla_r f_{l_{min}}(\hat{x})-\nabla_r f_{y}(\hat{x})\right)\cdot  \left(f_{y}(\hat{x})-f_{l_{min}}(\hat{x})\right) }{ \left \| \nabla_r f_{y}(\hat{x})-\nabla_r f_{l_{min}}(\hat{x}) \right \|_2^2}\)
}
return \((1 + \eta) \cdot r \)
\caption{Crossing \(k\) decision boundaries}
\label{alg:topkdeepfool} 
\end{algorithm}
\DecMargin{1em} 

\subsection{Multimodal Universal Perturbation}
Previously, we analyzed the method of obtaining the adversarial perturbation \(r\) for a sample point \(x_0\) in the linear scenario. 
% In this section, we will rely on the image-text retrieval task to design a novel algorithm in the multimodal scenario based on the above theoretical foundation, to generate universal adversarial perturbation for the VLP models.
In this section, based on the theoretical foundation outlined above, we will use the image-text retrieval task to design a novel algorithm for the multimodal scenario, generating a universal adversarial perturbation for VLP models.

\textbf{Problem Formulation.}
Let \(V = \{v_i\}_{i=1}^N\) and \(T = \{t_i\}_{i=1}^M\) represent an image set and a text set, respectively. 
\(E_v\) and \(E_t\) denote the visual encoder and text encoder, where \(E_v(v), E_t(t) \in \mathbb{R}^d\) represent the feature vectors normalized by \(\ell_2\) norm. 
We primarily consider two image-text downstream tasks: image-text retrieval and image classification. 
Taking the image-to-text retrieval task as an example, we define the indicator function \(I(E_v(v), E_t(T), k)\), where \(E_t(T) \in \mathbb{R}^{M \times d}\). 
First, we compute the cosine similarity between the visual embedding and the \(M\) text embeddings. 
If among the top \(k\) texts with the highest cosine similarity to \(E_v(v)\), there exists at least one text that matches \(v\), the value of the indicator function is 1; otherwise, it is 0.
Similarly, for text-to-image retrieval, the same indicator function is still applicable, denoted as \(I(E_t(t), E_v(V), k)\). 
% We try to find a universal perturbation \(\delta\) for the image modality which is sample-agnostic (i.e. the perturbation is shared by all images, rather than finding a perturbation for a specific image). 
We aim to identify a universal perturbation \(\delta\) for the image modality that is sample-agnostic, meaning it applies to all images instead of being specific to one.
The attack objective is as follows:
\begin{equation} \label{equ:reqv}
\forall v \in V, \,I(E_v(v+\delta),E(T),k)= 0 
\end{equation}
\begin{equation} \label{equ:reqt}
\forall t \in T, \,I(E_t(t),E(V+\delta),k)=0 
\end{equation}
where \(V + \delta = \{v_i + \delta\}_{i=1}^M\) represents adding the universal perturbation \(\delta\) to all images in the set \(V\).

For the image classification problem, we follow CLIP \cite{clip} to construct texts based on category names, forming the text set \(T = \{t_i\}_{i=1}^{C}\), where \(C\) is the number of categories, with texts like ``a photo of a xx.''
For an image, we calculate the cosine similarity between the image embedding and the text embeddings of all categories and classify the image into the most similar category.
For the classification task, perturbation \(\delta\) should satisfy:
\begin{equation}
\forall v \in V, \,
\underset{i}{argmax} 
( E_v(v+\delta)\cdot E_t(t_i) )  \neq y
\end{equation}
where \(y \in \{1, 2, \ldots, C\}\) is the true label of \(v\).

\IncMargin{1em}
\begin{algorithm}[t]
\SetKwData{Left}{left}\SetKwData{This}{this}\SetKwData{Up}{up} \SetKwFunction{Union}{Union}\SetKwInOut{Input}{Input}\SetKwInOut{Output}{Output}	
\Input{Image set \(V\), text set \(T\), mask matrix \(m\).} 
\Output{Perturbation \(\delta\).} 
Initialize \(\delta \gets \boldsymbol {0}\)
\BlankLine 
% \emph{special treatment of the first line}\; 
\For{\(v\) in \(V\)}
{
getting \(Y = \{y_j\}_{j=1}^{n}\) and \(Y' = \{y'_{j}\}_{j=1}^{k}\)

\(v \gets v \odot (1-m)+\delta \odot m\)

\(r \gets \boldsymbol {0}\)

\While{\(I(E_v(v + (1+\eta)\cdot r\odot m),E_t(T),k)==1\)}
{
\(\hat{v} \gets v + r\odot m \)

\(y'_{min} \gets \underset{y'\in Y'}{argmin} f_{y'}(\hat{v})\)

\(y_{max} \gets \underset{y\in Y}{argmax} f_{y}(\hat{v})\)

\(r \gets r + \frac{\left(\nabla_r f_{y'_{min}}(\hat{v})-\nabla_r f_{y_{max}}(\hat{v})\right)\cdot  \left(f_{y_{max}}(\hat{v})-f_{y'_{min}}(\hat{v})\right) }{ \left \| \nabla_r f_{y'_{min}}(\hat{v})-\nabla_r f_{y_{max}}(\hat{v}) \right \|_2^2}\)

}

\(\delta \gets clamp_{(0,1)}(\delta+(1+\eta)\cdot r)\)

}

return \( \delta \)
\caption{Attacking image-to-text retrieval}
\label{alg:attacki2t} 
\end{algorithm}
\DecMargin{1em} 

\textbf{Attacking image-to-text retrieval.}
We begin with a single image \(v \in \mathbb{R}^{c \times h \times w}\), where \(c\), \(h\), and \(w\) respectively stand for the number of channels, height, and width.
We denote \(Y = \{y_j\}_{j=1}^{n}\), where \(n\) represents that there are \(n\) texts in \(T\) that match \(v\), i.e. \(\{t_y\}_{y\in Y}\) match with \(v\). 
Let \(Y' = \{y'_{j}\}_{j=1}^{k}\), where \(Y'\) contains the indices of \(k\) texts that mismatch \(v\), and \(Y \cap Y' = \emptyset\). 
We consider the image-to-text retrieval task as a classification task, denoting \(f_i(v) = E_t(t_i) \cdot E_v(v)\), where \(E_t(t_i) \in \mathbb{R}^d\) plays the same role as \(\omega_i\) in the multi-class classifier. 
We will traverse the entire image set \(V\), but only optimize one perturbation \(\delta \in \mathbb{R}^{c \times h \times w}\). 
The pseudocode for generating \(\delta\) is shown in Alg~\ref{alg:attacki2t}. 
We must clarify four points: 
(1) \(Y\) can be directly obtained through the annotation information of the dataset.
(2) \(Y'\) serves as the target category for the attack. 
We can select the \(k\) nearest non-matching texts based on the decision boundary according to Equ~\ref{equ:l}. 
However, due to the large size of the text set, it will introduce a large amount of gradient backpropagation, severely affecting optimization efficiency. 
As an alternative, we use the cosine similarity score as the criterion and select the \(k\) non-matching texts with the highest similarity to \(E_v(v)\).
% (3) In line 11, \(P\) is a projection strategy that ensures \(\left \| \delta \right \| _p \le \epsilon\).
(3) 
% We believe that adding universal adversarial perturbation in patch form on images will have better effects compared to global perturbation which will be verified in subsequent experiments.
We hypothesize that applying universal adversarial perturbations in patch form to images will be more effective than global perturbations, a theory we will test in subsequent experiments.
Therefore, we introduce \(m \in \mathbb{R}^{c \times h \times w}\), a binary matrix that specifies the patch's location.
\(\odot\) denotes the element-wise product.
(4) \(clamp_{(0,1)}(\delta)\) represents clipping each element value in \(\delta\) to be between 0 and 1, obtaining valid pixel values.

\IncMargin{1em}
\begin{algorithm}[t]
\SetKwData{Left}{left}\SetKwData{This}{this}\SetKwData{Up}{up} \SetKwFunction{Union}{Union}\SetKwInOut{Input}{Input}\SetKwInOut{Output}{Output}	
\Input{Image set \(V\), text set \(T\), mask matrix \(m\).} 
\Output{Perturbation \(\delta\).} 
Initialize \(\delta \gets \boldsymbol {0}\)
\BlankLine 
% \emph{special treatment of the first line}\; 
\For{\(t\) in \(T\)}
{
getting \(y\) and \(Y' = \{y'_{j}\}_{j=1}^{k}\)

\(\tilde{V} = \{\tilde{v}_i\}_{i\in y\cup Y'} \gets \{v_i \odot (1-m)+\delta \odot m\}_{i\in y\cup Y'}\)

\(r \gets \boldsymbol {0}\)

\While{\(I(E_t(t),E_v(\tilde{V}+(1+\eta)\cdot r\odot m),k)==1\)}
{
\(\hat{V} =\{\hat{v}_i\}_{i\in y\cap Y'} \gets \tilde{V}+r\odot m\)

\(y'_{min} \gets \underset{y'\in Y'}{argmin}f_{y'}(t)\) \tcp{\(f_i(t)=E_v(\hat{v}_{i})\cdot E_t(t)\)}

\(r \gets r + \frac{\left(\nabla_r f_{y'_{min}}(t)-\nabla_r f_{y}(t)\right)\cdot  \left(f_{y}(t)-f_{y'_{min}}(t)\right) }{ \left \| \nabla_r f_{y'_{min}}(t)-\nabla_r f_{y}(t) \right \|_2^2}\)

}

\(\delta \gets clamp_{(0,1)}(\delta+(1+\eta)\cdot r)\)

}
return \( \delta \)
\caption{Attacking text-to-image retrieval}
\label{alg:attackt2i} 
\end{algorithm}
\DecMargin{1em} 

\textbf{Attacking text-to-image retrieval.} 
In this scenario, the most intuitive approach is to construct decision boundaries with \(E(V) \in \mathbb{R}^{N \times d}\) and generate adversarial text \(t_i^{adv}\) for text \(t_i\) to cross the decision boundary, similar to attacking image-to-text retrieval. 
However, due to the non-differentiability of textual data, generating the universal adversarial perturbation for texts is impractical.
% Therefore, we still generate the universal adversarial patch for the image modality, constructing malicious boundaries
Therefore, we continue to create the universal adversarial patch for the image modality, establishing malicious boundaries with \(E(V + \delta)\).
This implies that the text embedding remains unchanged, and when confronting the benign decision hyperplanes constituted by \(E(V)\), the text can accurately retrieve the matching image.
While facing the malicious decision boundaries formed by \(E(V + \delta)\), the text embedding is placed on an incorrect decision plane, leading to significant confusion.
The pseudocode is shown in Alg~\ref{alg:attackt2i}.
We need to clarify two points:
(1) A text matches only one image, with \(y\) representing the index of that image, obtained from the annotation information.
(2) \(Y'\) includes \(k\) non-matching images with the highest cosine similarity to \(E_v(t)\).

In summary, the inner loop concentrates on a single sample to determine the perturbation direction \(r\). 
In the outer loop, we continuously accumulate \(r\) onto \(\delta\), ultimately obtaining the universal adversarial perturbation applicable to all images.
Our method can be adapted to create universal adversarial global perturbations (Appendix C).
We don't specifically design the attack strategy for zero-shot classification task for two reasons: 
(1) Universal adversarial perturbation for the classification problem has been extensively studied in the unimodal field \cite{uap}.
(2) Our attack strategy demonstrates excellent transferability between datasets, meaning that \(\delta\) constructed based on the image-text retrieval task can be directly applied to classification datasets with good attack effectiveness.

\section{Experiments}

\begin{figure}[t]
\centering
\includegraphics[width=\linewidth]{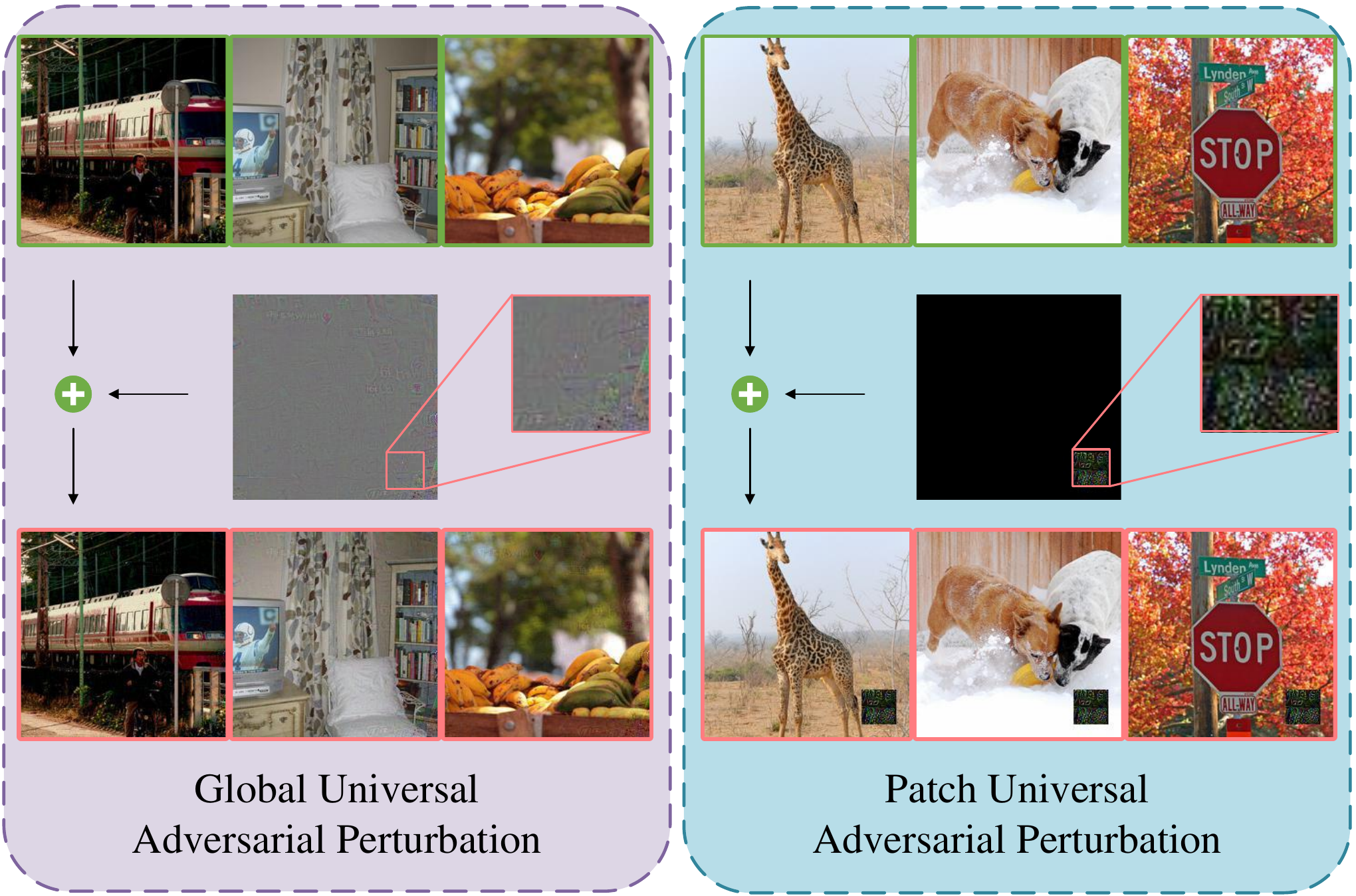}
\caption{
The top row shows the clean image, the middle row shows the universal adversarial perturbation (both global and patch), and the bottom row shows the adversarial image with the perturbation applied.}
\Description{The top row shows the clean image, the middle row shows the universal adversarial perturbation (both global and patch), and the bottom row shows the adversarial image with the perturbation applied.}
\label{tab:experience1}
\end{figure}

\begin{table}[t]
\caption{
The first column represents three different visual encoder configurations of the CLIP model. The pre-trained weights of CLIP are utilized directly, rather than weights fine-tuned on specific datasets.
}
\label{tab:clean}
\setlength{\tabcolsep}{3pt}
\begin{tabular}{ll|cccccc}
\toprule
\multirow{2}{*}{\parbox{1.2cm}{Visual \\Encoder}} & \multirow{2}{*}{Dataset} &\multicolumn{3}{c}{Test Retrieval} & \multicolumn{3}{c}{Image Retrieval} \\
\cmidrule(r){3-5} \cmidrule(r){6-8}
\multirow{2}{*}{}       &    & R@1   & R@5   & R@10  & R@1   & R@5   & R@10    \\
\midrule
\multirow{2}{*}{ViT-B/16}  & Flickr30k  & 82.20 & 96.70 & 98.90 & 62.12 & 85.72 & 91.94 \\
                           & COCO       & 52.40 & 76.78 & 84.66 & 33.07 & 58.35 & 69.03 \\
\midrule
\multirow{2}{*}{ViT-B/32}  & Flickr30k  & 78.60 & 94.90 & 98.30 & 58.80 & 83.56 & 90.00 \\
                           & COCO       & 50.18 & 75.02 & 83.50 & 30.48 & 55.99 & 66.87 \\
\midrule
\multirow{2}{*}{ViT-L/14}  & Flickr30k  & 85.30 & 97.40 & 99.20 & 64.84 & 87.22 & 92.04 \\
                           & COCO       & 56.34 & 79.40 & 86.58 & 36.51 & 61.08 & 71.17 \\
\bottomrule
\end{tabular}
\end{table}

In this section, we begin by presenting the performance of the CLIP model on the Flickr30K\cite{flickr30k} and MS COCO\cite{coco} datasets' test split for image-to-text retrieval task under the normal condition, in Tab~\ref{tab:clean}. 
This represents the baseline retrieval performance of the CLIP model. 
In an adversarial environment, the retrieval performance of CLIP will significantly deteriorate. 
The greater the decline, the more effective the attack.

\subsection{Exploring the Optimal Configuration}

\begin{table}[t]
\caption{The effects of universal patch attack with three different configurations.}
\label{tab:threeconfig}
\setlength{\tabcolsep}{3pt}
\begin{tabular}{ll|cccccc}
\toprule
\multirow{2}{*}{\parbox{1.2cm}{Visual \\Encoder}} & \multirow{2}{*}{Config} & \multicolumn{3}{l}{Text Retrieval} & \multicolumn{3}{l}{Image Retrieval} \\
\cmidrule(r){3-5} \cmidrule(r){6-8}
                                  &                         & R@1       & R@5       & R@10       & R@1       & R@5        & R@10       \\
\midrule
\multirow{3}{*}{ViT-B/16}         & TIRA                    & 0.10      & 0.60       & 1.00      & 0.06      & 0.36      & \textbf{0.54} \\
                                  & TRA                     & 1.60      & 4.00       & 7.40      & 25.92     & 51.24     & 62.36     \\
                                  & IRA                     & \textbf{0.00}    & \textbf{0.00}   & \textbf{0.20}  & \textbf{0.08} & \textbf{0.22} & 0.60      \\
\midrule
\multirow{3}{*}{ViT-B/32}         & TIRA                    & \textbf{0.50}    & \textbf{1.20}  & \textbf{2.50}  & 3.72      & 7.80      & 10.90     \\
                                  & TRA                     & 0.70      & 2.20       & 4.80      & 8.94      & 32.30     & 45.58     \\
                                  & IRA                     & 1.00      & 2.80       & 5.50      & \textbf{3.00}    & \textbf{6.44}  & \textbf{9.18}  \\
\midrule
\multirow{3}{*}{ViT-L/14}         & TIRA                    & 0.10      & 0.30       & 0.90      & 0.52      & 1.40      & 2.50      \\
                                  & TRA                     & 0.40      & 3.70       & 6.90      & 1.56      & 6.12      & 10.76     \\
                                  & IRA                     & \textbf{0.00}    & \textbf{0.00}   & \textbf{0.30}  & \textbf{0.10} & \textbf{0.22} & \textbf{0.36} \\
\bottomrule
\end{tabular}
\end{table}

In this section, we explore some configurations when generating universal adversarial patches based on the CLIP$_{vit/B16}$, CLIP$_{vit/B32}$ and CLIP$_{vit/L14}$ models. 
We use the first 1000 images from the Flickr30k training set and the corresponding 5000 texts to construct \(V=\{v_i\}_{i=1}^{1000}\) and \(T=\{t_i\}_{i=1}^{5000}\), meaning that only 1000 images are visible during the optimization process of generating the universal adversarial patch.
we use the mask matrix \(m\) to ensure that the patch area is 3\% of the total image area. 
Since the input size of the CLIP model is \(224\times 224\), the patch size is \(38\times 38\), and it is fixed in the bottom right corner of the image, as shown in Fig~\ref{tab:experience1}.
We set \(k\) to 10, as we will evaluate the R@10 metric.
During the evaluation, We apply the universal patch to all images in the Flickr30k test split and conduct the image-text retrieval task using the CLIP model. 

In Alg~\ref{alg:attacki2t} and Alg~\ref{alg:attackt2i}, we have provided detailed descriptions of how to attack the text retrieval and image retrieval tasks. 
These two attack strategies can be used independently (denoted as \textbf{IRA} and \textbf{TRA} for short), or in conjunction (``Text retrieval and Image Retrieval Attack'' is denoted as \textbf{TIRA}). 
See Appendix A for details.
We conduct a comparison of three configurations based on patch form.
The results are shown in Tab~\ref{tab:threeconfig}, and we find that using TRA alone results in poor attack performance on the image retrieval (i.e., text-to-image retrieval) sub-task. 
However, employing IRA alone can achieve good attack performance on both sub-tasks.
During model training, one image corresponds to multiple texts, while one text corresponds to only a single image. 
Besides the cross-modal contrastive representation learning between images and texts, the textual modality implicitly includes within-modality contrastive representation clustering, which does not exist in the visual modality. 
When using TRA to drive image representations away from similar texts, it does not disrupt the topological structure among the image representations. 
That is, from the perspective of text representations, all matching image representations are moving away, but the similarity ranking remains largely unchanged. 
Therefore, the performance decline in the image retrieval sub-task is minimal. 
Based on this, we speculate that introducing contrastive learning between images during the visual-language pre-training process might enhance model robustness, a hypothesis that awaits exploration in future work.

\begin{figure}[t]
  \centering
  \includegraphics[width=\linewidth]{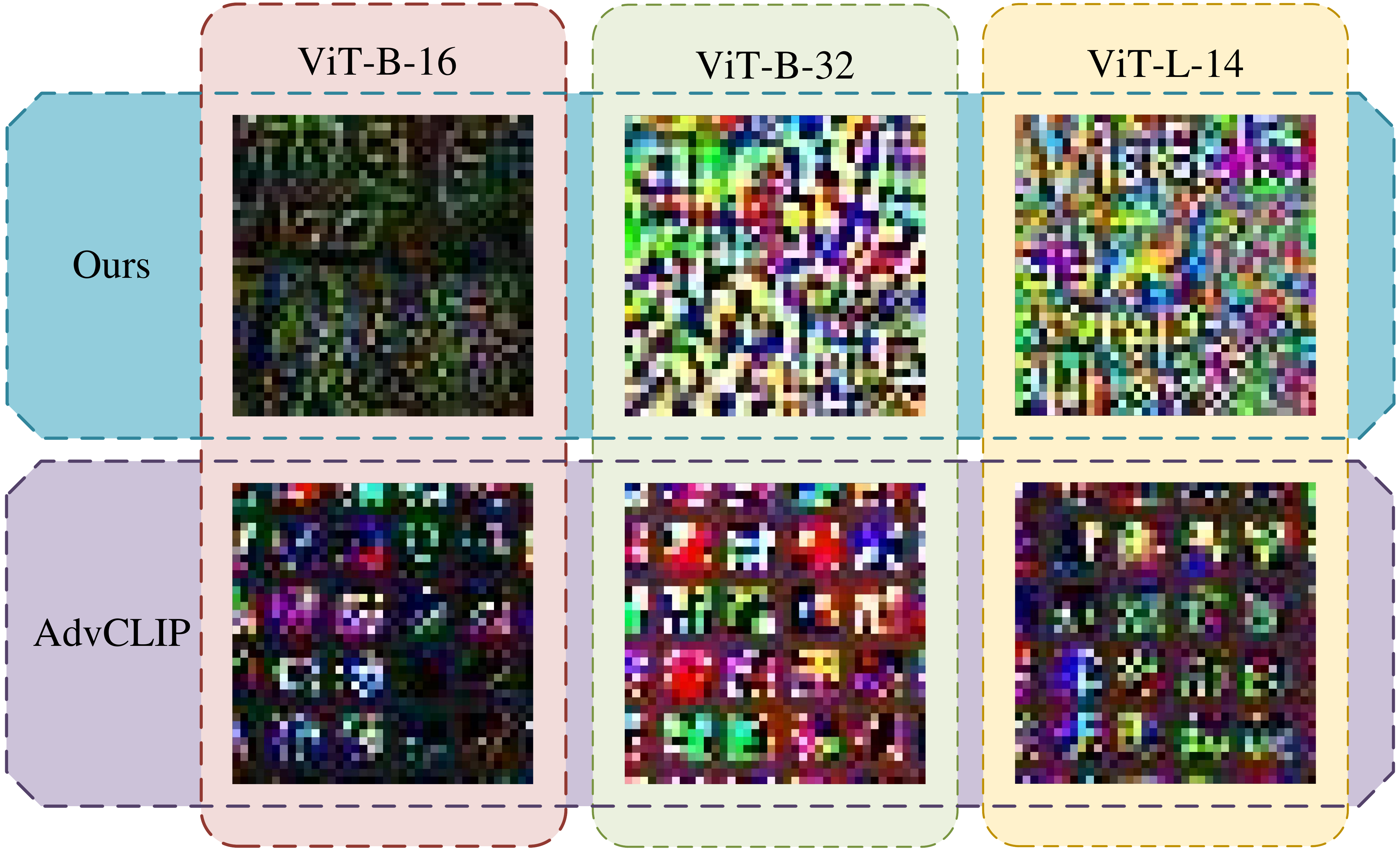}
  \caption{
    Visualization of universal adversarial patches generated by two methods on three models.}
  \label{fig:sixpatches}
  \Description{Universal adversarial patches generated by two methods on three models.}
\end{figure}

\begin{table*}[t]
\caption{Comparing between AdvCLIP and UAP\(^{patch}_{TIRA}\), based on the Flickr30k and MS COCO datasets. Recall rates are reported.}
\label{tab:advclip1}
\begin{tabular}{ll|ccc|ccc|ccc|ccc}
\toprule
\multirow{3}{*}{\parbox{1.2cm}{Visual \\Encoder}} & \multirow{3}{*}{Method} & \multicolumn{6}{c|}{Flickr30k(1K test set)}                         & \multicolumn{6}{c}{MS COCO(5K test set)}                        \\
\multicolumn{2}{c|}{}                        & \multicolumn{3}{c}{TR} & \multicolumn{3}{c|}{IR} & \multicolumn{3}{c}{TR} & \multicolumn{3}{c}{IR} \\
\cmidrule(r){3-5} \cmidrule(r){6-8} \cmidrule(r){9-11} \cmidrule(r){12-14} 
\multicolumn{2}{c|}{}                        & R@1   & R@5    & R@10     & R@1   & R@5    & R@10     & R@1   & R@5    & R@10     & R@1   & R@5    & R@10     \\
\midrule
\multirow{2}{*}{ViT-B/16} & Advclip          & 2.30  & 7.00    & 11.20    & 19.98 & 48.08 & 64.38     & 1.58  & 4.44  & 6.54      & 8.10   & 23.55 & 34.49     \\
                          & UAP\(^{patch}_{TIRA}\) & \textbf{0.10}  & \textbf{0.60}  & \textbf{1.00}    & \textbf{0.06}  & \textbf{0.36}  & \textbf{0.54}    & \textbf{0.02}   & \textbf{0.04}  & \textbf{0.10}   & \textbf{0.18}  & \textbf{0.51}  & \textbf{0.75}    \\
\midrule
\multirow{2}{*}{ViT-B/32} & Advclip          & 10.80 & 23.40 & 32.80     & 25.68  & 54.94 & 67.60      & 7.18  & 17.77 & 24.34     & 11.66 & 29.50  & 40.84     \\
                          & UAP\(^{patch}_{TIRA}\) & \textbf{0.50}  & \textbf{1.20}  & \textbf{2.50}    & \textbf{3.72}  & \textbf{7.80}   & \textbf{10.90}    & \textbf{0.38}     & \textbf{1.30}   & \textbf{2.00}   & \textbf{2.00}   & \textbf{5.35}  & \textbf{8.20}     \\
\midrule
\multirow{2}{*}{ViT-L/14} & Advclip          & 4.80  & 14.40 & 19.20     & 18.28 & 44.18 & 55.82      & 0.84  & 3.18  & 5.06      & 6.84  & 19.32 & 28.87     \\
                          & UAP\(^{patch}_{TIRA}\) & \textbf{0.10}  & \textbf{0.30}  & \textbf{0.90}    & \textbf{0.52}   & \textbf{1.40}   & \textbf{2.50}    & \textbf{0.12}  & \textbf{0.20}   & \textbf{0.36} & \textbf{0.32} & \textbf{0.77}  & \textbf{1.14} \\
\bottomrule
\end{tabular}
\end{table*}

\subsection{Comparing with SOTA}

In this section, we compare our method with AdvCLIP\cite{advclip}, which is currently the best method for generating universal adversarial patches against VLP models. 
We denote our proposed universal patch attack (using TIRA) as UAP\(^{patch}_{TIRA}\). See Appendix B for more experiments about UAP\(^{patch}_{IRA}\).
For fairness, both approaches utilize the same 1000 images from Flickr30K as the surrogate images and maintain identical patch size and position mentioned above. 
Based on three CLIP models, the six universal adversarial patches generated by two methods are shown in Fig~\ref{fig:sixpatches}.

\textbf{Image-text Retrieval.} 
The effects of two attack methods are presented in Tab~\ref{tab:advclip1}. 
We observe that: 
(1) The universal adversarial patches generated by the Advclip perform poorly in the image retrieval sub-task. 
(2) Our patches outperform Advclip across all metrics. 
(3) Our patches, generated based on Flickr30K images demonstrate strong attack effectiveness even when applied to the MS COCO dataset, indicating that our method possesses good \textbf{cross-dataset transferability}.

\begin{table}[t]
\caption{Comparing between AdvCLIP and UAP\(^{patch}_{TIRA}\), based on the ImageNet, CIFAR-100 and CIFAR-10 datasets. The accuracy is reported. Lower is better.}
\label{tab:advclip2}
\setlength{\tabcolsep}{3pt}
\renewcommand\arraystretch{1.2}
\begin{tabular}{ll|cc|cc|cc}
\toprule
\multirow{2}{*}{\parbox{1.2cm}{Visual \\Encoder}} & \multirow{2}{*}{Method} & \multicolumn{2}{c|}{ImageNet}    & \multicolumn{2}{c|}{CIFAR100}    & \multicolumn{2}{c}{CIFAR10}                    \\
\cmidrule(r){3-4} \cmidrule(r){5-6} \cmidrule(r){7-8}  
\multicolumn{2}{c|}{}     & Top1     & Top5    & Top1   &Top5   & Top1   & Top5  \\
\midrule
\multirow{3}{*}{ViT-B/16} & w/o atk & 62.42 & 86.74 & 66.56 & 88.49 & 90.10 & 99.07 \\
                        & Advclip & 5.98  & 14.80 & 8.91  & 45.04 & 81.90 & 97.01 \\
                        &  UAP\(^{patch}_{TIRA}\)     & \textbf{0.12}  &\textbf{ 0.48}  & \textbf{1.62}  & \textbf{5.82}  & \textbf{10.67} & \textbf{53.04} \\
\midrule                   
\multirow{3}{*}{ViT-B/32} & w/o atk & 57.50 & 83.57 & 62.27 & 86.98 & 88.34 & 99.24 \\
                        & Advclip & 7.98  & 27.40 & 31.92 & 51.84 & 64.45 & 96.51 \\
                        &  UAP\(^{patch}_{TIRA}\) & \textbf{1.68}  & \textbf{5.36}  & \textbf{7.22}  & \textbf{17.76} & \textbf{15.10} & \textbf{68.69} \\
\midrule
\multirow{3}{*}{ViT-L/14} & w/o atk & 69.74 & 90.15 & 75.72 & 93.06 & 95.19 & 99.52 \\
                        & Advclip & 1.54  & 7.61  & 7.42  & 32.30 & 73.87 & 97.95 \\
                        &  UAP\(^{patch}_{TIRA}\) & \textbf{0.58}  & \textbf{1.61 } & \textbf{6.57}  & \textbf{15.44} &\textbf{ 20.76 }& \textbf{70.55} \\              
\bottomrule
\end{tabular}
\end{table}

\begin{figure}[t]
\centering
\includegraphics[width=\linewidth]{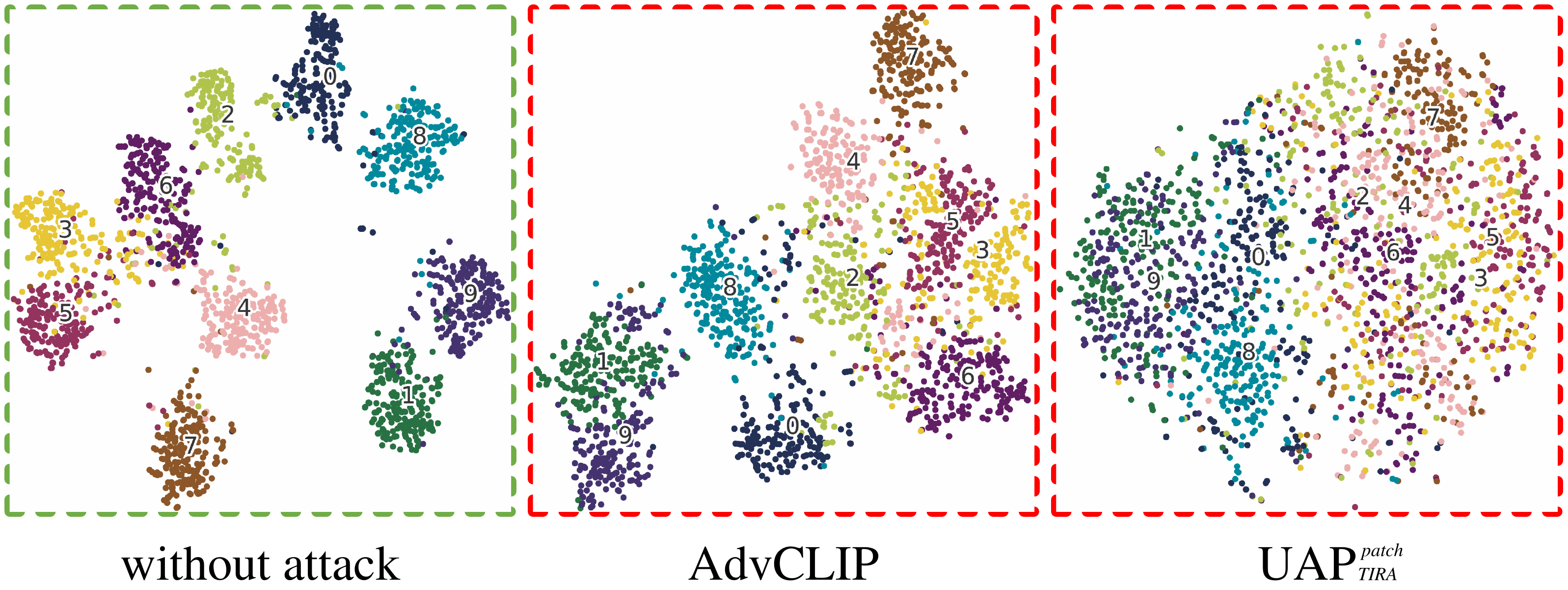}
\caption{
Based on the CIFAR-10 dataset and the CLIP\(_{vit/B16}\) model, t-SNE feature dimensionality reduction visualization reveals that upon the addition of the universal adversarial patch generated by UAP\(^{patch}_{TIRA}\), the features extracted by the visual encoder become severely disordered, leading directly to a collapse in model performance.}
\Description{Based on the CIFAR-10 dataset and the CLIP\(_{ViT-B/16}\) model, t-SNE feature dimensionality reduction visualization reveals that upon the addition of the universal adversarial patch generated by UAP\(^{patch}_{TIRA}\), the features extracted by the visual encoder become severely disordered, leading directly to a collapse in model performance.}
\label{fig:t-sne}
\end{figure}

\textbf{Zero-shot  Classification.} 
We perform this task on three commonly used classification datasets: ImageNet1k\cite{imagenet}, CIFAR-100\cite{cifar10}, and CIFAR-10\cite{cifar10}. 
It's important to note that we continue to use the six universal adversarial patches from Fig~\ref{fig:sixpatches}, instead of generating unique patches for each dataset. 
We resize the images to \(224\times 224\) to meet the requirement of the CLIP models.
The effects of the attacks are presented in Tab~\ref{tab:advclip2}. 
It can be observed that: 
(1) Our patches consistently outperform Advclip in terms of attack effectiveness.
(2) The dataset with fewer categories is less susceptible to attack because the fewer the categories, the sparser the decision boundaries, making samples farther from these boundaries and crossing them more difficult. 
(3) Our patches exhibit good \textbf{cross-task transferability}.
We employ t-SNE\cite{tsne} to reduce the dimensionality and visualize features extracted by the ViT-B/16 visual encoder on the CIFAR-10 dataset, as shown in Fig~\ref{fig:t-sne}. 
Our method evidently results in a more chaotic feature distribution.

\begin{table*}[t]
\caption{Comparing between AdvCLIP and UAP\(^{patch}_{TIRA}\), based on the NUS-WIDE, Pascal-Sentence and Wikipedia datasets.}
\label{tab:advclip3}
\begin{tabular}{ll|ccccccccc}
\toprule
\multirow{2}{*}{Encoder} & \multirow{2}{*}{Metric} & \multicolumn{3}{c}{NUS-WIDE} & \multicolumn{3}{c}{Pascal-Sentence} & \multicolumn{3}{c}{Wikipedia} \\
\cmidrule(r){3-5} \cmidrule(r){6-8} \cmidrule(r){9-11}  
                         &                         & w/o atk  & AdvCLIP  & UAP\(^{patch}_{TIRA}\)   & w/o atk  & AdvCLIP & UAP\(^{patch}_{TIRA}\)  & w/o atk   & AdvCLIP  & UAP\(^{patch}_{TIRA}\)   \\
\midrule
\multirow{5}{*}{ViT-B/16}  & TR@1                    & 66.90    & 36.75    & \textbf{11.50}  & 67.00    & 7.00    & \textbf{5.00}  & 61.69     & 14.72    & \textbf{6.93}   \\
                         & IR@1                    & 61.80    & 51.25    & \textbf{16.80}  & 68.50    & 10.00   & \textbf{7.50}  & 66.67     & 42.64    & \textbf{23.16}  \\
                         & AVG                     & 64.35    & 44.00    & \textbf{14.15}   & 67.75    & 8.50    & \textbf{6.25}  & 64.18     & 28.68    & \textbf{15.05}  \\
                         & Top1                    & 77.85    & 49.65    & \textbf{11.25}  & 77.00    &  5.50   & \textbf{5.50}  & 64.01     & 27.95    & \textbf{7.09}   \\
                         & Top5                    & 98.75    & 72.95    & \textbf{47.55}  & 96.00    & 39.50   & \textbf{29.00} & 94.35     & 73.15    & \textbf{52.75}  \\
\midrule
\multirow{5}{*}{ViT-B/32}  & TR@1                    & 69.25    & 34.15    & \textbf{23.25}  & 72.00    & 35.00   & \textbf{10.50} & 58.23     & 31.60    & \textbf{17.97}  \\
                         & IR@1                    & 60.25    & 47.60    & \textbf{39.10}  & 70.50    & 47.00   & \textbf{17.00} & 62.77     & 48.70    & \textbf{18.61}  \\
                         & AVG                     & 64.75    & 40.88   & \textbf{31.18}  & 71.25    & 41.00   & \textbf{13.75} & 60.5      & 40.15    & \textbf{18.29}  \\
                         & Top1                    & 77.65    & 32.60    & \textbf{28.15}  & 72.00    & 52.00   & \textbf{9.00} & 62.09     & 35.33    & \textbf{19.87}  \\
                         & Top5                    & 98.55    & 87.05    & \textbf{66.00}  & 94.50    & 88.00   & \textbf{42.00} & 93.91     & 81.30    & \textbf{62.10}  \\
\midrule
\multirow{5}{*}{ViT-L/14}  & TR@1                    & 67.50    & 37.15    & \textbf{12.80}  & 65.50    & 17.00   & \textbf{4.00}  & 62.55     & 40.48    & \textbf{19.48}  \\
                         & IR@1                    & 61.15    & 50.10    & \textbf{48.55}  & 68.00    & 9.50    & \textbf{2.00}  & 62.77     & 50.43    & \textbf{33.98}  \\
                         & AVG                     & 64.325   & 43.625   & \textbf{30.68}  & 66.75    & 13.25   & \textbf{3.00}  & 62.66     & 45.455   & \textbf{26.73}  \\
                         & Top1                    & 76.35    & 16.15    & \textbf{11.30}  & 75.00    & 22.00   & \textbf{5.00}  & 67.09     & 35.36    & \textbf{17.99}   \\
                         & Top5                    & 98.20    & 76.05    & \textbf{57.55}  & 94.20    & 69.50   & \textbf{26.00} & 94.23     & 74.63    & \textbf{52.25}  \\
\bottomrule
\end{tabular}
\end{table*}

\textbf{Evaluated on Fine-tuned Models.} 
We select three datasets, namely NUS-WIDE\cite{nus-wide}, Pascal-Sentence\cite{pascal-sentence}, and Wikipedia\cite{wikipedia}, that support both image-text retrieval and image classification. 
The data format is a triplet \((v, t, y)\), consisting of matching image \(v\) and text \(t\) with their category \(y\).
Based on the three datasets, two downstream tasks, and three CLIP pre-trained models, we obtained a total of 18 fine-tuned models.
Considering the powerful zero-shot capability of the CLIP model, we follow the setup of AdvCLIP\cite{advclip}, freezing the weights of CLIP encoders and only fine-tuning the subsequent nonlinear layers.
For the image-text retrieval, we add a nonlinear projection head after both the text and visual encoders respectively to obtain feature vectors, and then perform classification supervised by category annotation information through the same linear layer. 
During the testing phase, we discard the linear classification head and calculate the cosine similarity between embeddings output by the nonlinear projection heads to perform the image-text retrieval task. 
We only measure the R@1 metric, with text retrieval R@1 and image retrieval R@1 abbreviated as TR@1 and IR@1, respectively.
For the image classification, we measure the Top1 and Top5 accuracy.
Note, the adversarial patches remain the six patches from Fig~\ref{fig:sixpatches}, aimed at observing the attack effectiveness of perturbations generated based on pre-trained models when transferred to fine-tuned models. 
From Tab~\ref{tab:advclip3} we observed that our method consistently maintains better and more stable attack performance, demonstrating its excellent \textbf{cross-model transferability}. 
Due to the shared parameters between the encoder of the pre-trained model and the encoder of the fine-tuned model, it does not constitute a transfer attack in the absolute sense.

We recognize that our approach requiring precise perturbation distance and direction determination, involves extracting an individual adversarial direction for each image and obtaining a universal perturbation ultimately.
As a result, our method consumes more optimization time than AdvCLIP.
But these additional time overheads occur before the perturbation is deployed, not affecting the real-time nature of the attack strategy, and are therefore entirely acceptable.
Due to the absence of batch processing, our method's effectiveness is not influenced by batch size, in contrast to AdvCLIP, which exhibits significant performance fluctuation under different batch-size values according to its original paper\cite{advclip}.
Furthermore, our method requires few GPU computational resources. 
For instance, utilizing the CLIP model to iterate over 1,000 Flickr30k surrogate images 50 times on a single NVIDIA RTX4090 GPU necessitates 2430MB of memory and approximately 18 hours of optimization time.

\begin{table}[t]
\caption{The effects of universal adversarial patches on BEiT3.}
\label{tab:patchbeit3}
\setlength{\tabcolsep}{3.8pt}
\begin{tabular}{lll|ccc|ccc}
\toprule
\multirow{2}{*}{MD} & \multirow{2}{*}{DS} & \multirow{2}{*}{Patch}  & \multicolumn{3}{c}{Text Retrieval}  & \multicolumn{3}{c}{Image Retrieval}    \\
\cmidrule(r){4-6} \cmidrule(r){7-9} 
\multicolumn{3}{c|}{}                        & R@1     & R@5    & R@10   & R@1     & R@5    & R@10   \\
\midrule
\multirow{6}{*}{\(M1\)}    & \multirow{3}{*}{\(D1\)}      & w/o atk     & 78.98 & 94.38 & 97.20  & 61.36 & 84.61 & 90.74 \\
                           &                            & \(Patch1\)  & 0.22  & 0.72  & 1.12   & 1.66  & 4.72  & 6.74  \\
                           &                            & \(Patch2\)  & 0.20  & 0.58  & 1.00   & 2.12  & 5.29  & 7.53  \\
\cmidrule(r){2-9}
                           & \multirow{3}{*}{\(D2\)} & w/o atk   & 93.60 & 99.30 & 99.80  & 82.90 & 96.54 & 98.46 \\
                           &                            & \(Patch1\)  & 0.70  & 2.70  & 4.10   & 5.42  & 12.42 & 17.10 \\
                           &                            & \(Patch2\)  & 0.40  & 1.00  & 2.00   & 1.92  & 4.64  & 7.12  \\
\midrule
\multirow{6}{*}{\(M2\)}    & \multirow{3}{*}{\(D1\)}      & w/o atk     & 71.00 & 90.16 & 94.44  & 52.79 & 77.12 & 85.07 \\
                           &                            &\(Patch1\)   & 1.74  & 4.48  & 6.20   & 5.61  & 13.97 & 19.38 \\
                           &                            & \(Patch2\)  & 0.86  & 2.76  & 4.46   & 5.10  & 13.40 & 19.40 \\
\cmidrule(r){2-9}
                           & \multirow{3}{*}{\(D2\)} & w/o atk     & 96.30 & 99.70 & 100.00 & 86.14 & 97.68 & 98.82 \\
                           &                            & \(Patch1\)  & 4.10  & 12.10 & 16.90  & 17.92 & 37.02 & 46.80 \\
                           &                            & \(Patch2\)  & 1.70  & 3.80  & 6.10   & 11.58 & 24.64 & 32.06 \\
\bottomrule
\end{tabular}
\end{table}

\subsection{Extend to Larger VLP models}
Considering the CLIP pre-trained models do not exhibit state-of-the-art performance in the non-adversarial environment, we chose the recently open-sourced VLP model BEiT3\cite{beit3}, which has a larger parameter size and more advanced image-text interactive capabilities, to continue validating the effectiveness of our attack method.
\(D1\) and \(D2\) represent the MS COCO and Flickr30k datasets. 
\(M1\) and \(M2\) denote the BEiT3 models fine-tuned on D1 and D2, respectively. 
\(Patch1\) and \(Patch2\) are universal adversarial patches generated based on \(M1\) using 1000 training images from MS COCO and 1000 training images from Flickr30k, respectively. 
\(M2\) is reserved for testing the cross-model transferability, with experimental results shown in Tab~\ref{tab:patchbeit3}.
The experimental results validate that our universal adversarial patches possess stable and potent attack effectiveness, as well as good cross-dataset and cross-model transferability.
See Appendix D for more experiments on BEiT3.

\subsection{Global Perturbation}

\begin{figure}[t]
\centering
\includegraphics[width=\linewidth]{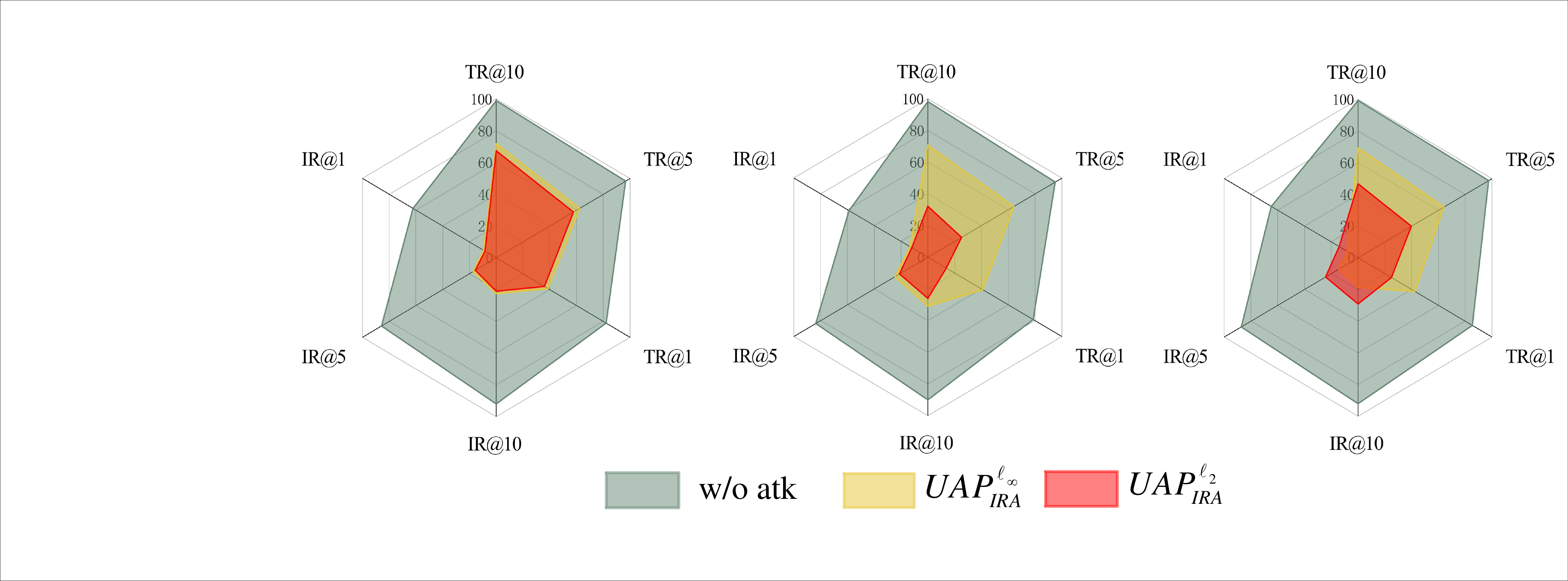}
\caption{Universal adversarial global perturbation results based on Flickr30k. From left to right, we utilize CLIP$_{vit/B16}$, CLIP$_{vit/B32}$, and CLIP$_{vit/L14}$, respectively.}
\label{fig:global-flickr}
\Description{Global universal adversarial perturbation results based on Flickr30k. From left to right, we utilize CLIP$_{ViT\-B\/16}$, CLIP$_{ViT\-B\/32}$, and CLIP$_{ViT\-L\/14}$, respectively.}
\end{figure}

\begin{figure}[t]
\centering
\includegraphics[width=\linewidth]{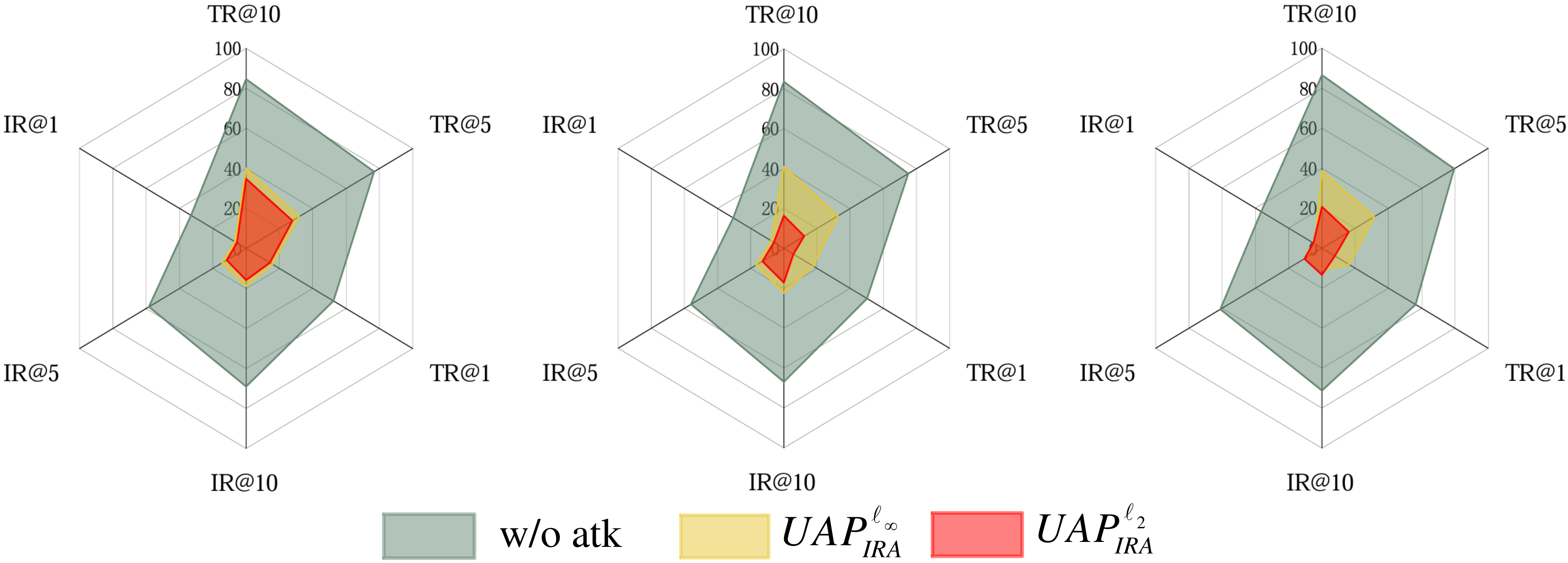}
\caption{Universal adversarial global perturbation results based on MS COCO, for verifying cross-dataset transferability.}
\Description{Global universal adversarial perturbation results based on MS COCO, for verifying cross-dataset transferability.}
\label{fig:global-coco}
\end{figure}

Our method can also generate universal adversarial perturbations in global form. 
See Appendix C for technical details. 
We use 1000 Flickr30k images as proxy images, utilizing only the IRA (Image Retrieval Attack) under constraints based on the \(\ell_2\) norm (\(\epsilon_{\ell_2}=2000\)) and \(\ell_{\infty}\) norm (\(\epsilon_{\ell_{\infty}}=10\)), following \cite{uap}.
Fig~\ref{fig:global-flickr} shows the performance degradation of the CLIP models when universal global perturbations are applied to the entire Flickr30k test dataset. 
Fig~\ref{fig:global-coco} shows the results of transferring global perturbations to the COCO dataset to demonstrate their cross-dataset transferability.
We observe that under \(\ell_2\) norm constraint, universal global perturbations generally achieve better attack effectiveness.
Visualization of global perturbations can be found in Appendix C.

\section{CONCLUSION}
% To the best of our knowledge, this paper is the first to apply decision boundary theory to the multimodal scenario, providing a theoretical foundation for exploring the robustness of multimodal embeddings within the high-dimensional space.
To the best of our knowledge, this is the first work to apply decision boundary theory to the multimodal scenario. 
We provide a theoretical foundation for exploring the robustness of multimodal embeddings within the high-dimensional space.
Based on this, a new method is proposed to improve understanding of VLP models through the development of sample-agnostic perturbations.
Our method differs from sample-specific perturbation techniques by using decision boundaries to create universal perturbations.
% Our approach diverges from sample-specific perturbation methods by utilizing decision boundaries to generate universal perturbations. 
% It explores a unique interaction between visual and textual modalities, treating one as the decision boundary for the other, and vice versa. 
It investigates a unique interaction between visual and textual modalities, with each serving as the decision boundary for the other.
We further validated the universality of the perturbation based on larger models and the global perturbation form.
Experimental results demonstrate promising transferability across images, tasks, and models, providing a scalable and efficient strategy to explore the security of multimedia and multimodal technologies.
% The experimental results show promising transferability across images, tasks, and models, offering a more scalable and efficient strategy for understanding and exploring the security of multimedia and multimodal processing technologies.
We hope our code can be beneficial to the community.

\appendix

\section{Combining IRA and TRA} \label{Combining IRA and TRA}
% √

%%%%%%%%%%%%%%%%%%%%%%%%%%%%%%%%%%%%%%%%%%%%%%%%%%%%%%%%%%%%%%%%%%%%%%%%
% 算法TIRA开始
%%%%%%%%%%%%%%%%%%%%%%%%%%%%%%%%%%%%%%%%%%%%%%%%%%%%%%%%%%%%%%%%%%%%%%%%
\IncMargin{1em}
\begin{algorithm}[]
\SetKwData{Left}{left}\SetKwData{This}{this}\SetKwData{Up}{up} \SetKwFunction{Union}{Union}\SetKwInOut{Input}{Input}\SetKwInOut{Output}{Output}	
\Input{Image set \(V\), text set \(T\), mask matrix \(m\).} 
\Output{Perturbation \(\delta\).} 
Initialize \(\delta \gets \boldsymbol {0}\)
\BlankLine 
% \emph{special treatment of the first line}\; 
\tcp{getting a batch of images each time \(\tilde{V}\)}
% \tcp{Get matching text set \(\tilde{T}\) of \(\tilde{V}\)}
\For{\(\tilde{V}\) in \(V\)} 
{
get matching text set \(\tilde{T}\) of \(\tilde{V}\)

\(r \gets \boldsymbol {0}\)

\For{\(v\) in \(\tilde{V}\)} 
{traverse \(\tilde{V}\) to execute IRA}

\(\delta \gets clamp_{(0,1)}(\delta+(1+\eta)\cdot r)\)

\(r \gets \boldsymbol {0}\)

\For{\(t\) in \(\tilde{T}\)} 
{traverse \(\tilde{T}\) to execute TRA}

\(\delta \gets clamp_{(0,1)}(\delta+(1+\eta)\cdot r)\)

}
return \( \delta \)
\caption{TIRA (combining IRA and TRA)}
\label{alg:tira} 
\end{algorithm}
\DecMargin{1em} 
%%%%%%%%%%%%%%%%%%%%%%%%%%%%%%%%%%%%%%%%%%%%%%%%%%%%%%%%%%%%%%%%%%%%%%%%
% 算法TIRA结束
%%%%%%%%%%%%%%%%%%%%%%%%%%%%%%%%%%%%%%%%%%%%%%%%%%%%%%%%%%%%%%%%%%%%%%%%
% 还需品一品

Image Retrieval universal perturbation Attack (IRA) and Text Retrieval universal perturbation Attack (TRA) can be used independently or in combination (TIRA), as shown in Alg~\ref{alg:tira}.
% √
During TIRA, for each iteration, a subset \(\tilde{V}\) is extracted from the image set \(V\). 
% √
Concurrently, based on annotation information, all texts that match \(\tilde{V}\) are collected to construct the text subset \(\tilde{T}\).
% √
IRA and TRA are executed separately based on \(\tilde{V}\) and \(\tilde{T}\).
% √
During IRA, benign decision boundaries are constructed with \(E_t(\tilde{T})\), and the optimization goal is to use universal adversarial perturbation to push \(v \in \tilde{V}\) across these decision boundaries, compromising the performance of text retrieval.
% √
During TRA, malicious boundaries are established with \(E_t(\tilde{V'})\) with the text input remaining unchanged.
% √
\(\tilde{V'}\) represents the subset of images where each \(v \in \tilde{V}\) has been modified with the universal adversarial perturbation.
% √

%%%%%%%%%%%%%%%%%%%%%%%%%%%%%%%%%%%%%%%%%%%%%%%%%%%%%%%%%%%%%%%%%%%%%%%%
% 表格对比Advclip与UAP IRA patch开始 MS COCO FLICKR30K
%%%%%%%%%%%%%%%%%%%%%%%%%%%%%%%%%%%%%%%%%%%%%%%%%%%%%%%%%%%%%%%%%%%%%%%%
\begin{table*}[t]
\caption{Comparing between AdvCLIP and UAP\(^{patch}_{IRA}\), based on the Flickr30k and MS COCO datasets, the recall rate is reported. Lower is better.}
\label{tab:advclip1_}
\renewcommand\arraystretch{1.2}
\begin{tabular}{ll|ccc|ccc|ccc|ccc}
\toprule
\multirow{3}{*}{\parbox{1.2cm}{Visual \\Encoder}} & \multirow{3}{*}{Method} & \multicolumn{6}{c|}{Flickr30k(1K test set)}                         & \multicolumn{6}{c}{MS COCO(5K test set)}                        \\
\multicolumn{2}{c|}{}                        & \multicolumn{3}{c}{TR} & \multicolumn{3}{c|}{IR} & \multicolumn{3}{c}{TR} & \multicolumn{3}{c}{IR} \\
\cmidrule(r){3-5} \cmidrule(r){6-8} \cmidrule(r){9-11} \cmidrule(r){12-14} 
\multicolumn{2}{c|}{}                        & R@1   & R@5    & R@10     & R@1   & R@5    & R@10     & R@1   & R@5    & R@10     & R@1   & R@5    & R@10     \\
\midrule
\multirow{2}{*}{ViT-B/16} & AdvCLIP & 2.30  & 7.00    & 11.20 & 19.98 & 48.08 & 64.38 & 1.58 & 4.44  & 6.54  & 8.10   & 23.55 & 34.49 \\
                          & UAP\(^{patch}_{IRA}\)& \textbf{0.00}  & \textbf{0.00}  & \textbf{0.20}  & \textbf{0.08}  & \textbf{0.22}  & \textbf{0.60}  & \textbf{0.00}  & \textbf{0.10}  & \textbf{0.16}   & \textbf{0.18}  & \textbf{0.37}  & \textbf{0.55}  \\
\midrule
\multirow{2}{*}{ViT-B/32} & AdvCLIP & 10.80 & 23.40 & 32.80  & 25.68 & 54.94 & 67.60  & 7.18 & 17.77 & 24.34 & 11.66 & 29.50  & 40.84 \\
                          &UAP\(^{patch}_{IRA}\) & \textbf{1.00}  & \textbf{2.80}  & \textbf{5.50}  & \textbf{3.00}  & \textbf{6.44}  & \textbf{9.18}  & \textbf{0.60}  & \textbf{2.34}  & \textbf{3.74}  & \textbf{2.39}   & \textbf{6.02}   & \textbf{8.83}   \\
\midrule
\multirow{2}{*}{ViT-L/14} & AdvCLIP & 4.80  & 14.40 & 19.20  & 18.28 & 44.18 & 55.82  & 0.84 & 3.18  & 5.06  & 6.84  & 19.32 & 28.87 \\
                          & UAP\(^{patch}_{IRA}\) & \textbf{0.00}  & \textbf{0.00}  & \textbf{0.30}  & \textbf{0.10}  & \textbf{0.22}  & \textbf{0.36}  & \textbf{0.06}  & \textbf{0.08}  & \textbf{0.14}  & \textbf{0.11}  & \textbf{0.27}  & \textbf{0.39} \\
\bottomrule
\end{tabular}
\end{table*}
%%%%%%%%%%%%%%%%%%%%%%%%%%%%%%%%%%%%%%%%%%%%%%%%%%%%%%%%%%%%%%%%%%%%%%%%
% 表格对比Advclip与UAP IRA patch结束 MS COCO FLICKR30K
%%%%%%%%%%%%%%%%%%%%%%%%%%%%%%%%%%%%%%%%%%%%%%%%%%%%%%%%%%%%%%%%%%%%%%%%
% √

%%%%%%%%%%%%%%%%%%%%%%%%%%%%%%%%%%%%%%%%%%%%%%%%%%%%%%%%%%%%%%%%%%%%%%%%
% TRA global版 开始
%%%%%%%%%%%%%%%%%%%%%%%%%%%%%%%%%%%%%%%%%%%%%%%%%%%%%%%%%%%%%%%%%%%%%%%%
\IncMargin{1em}
\begin{algorithm}[b]
\SetKwData{Left}{left}\SetKwData{This}{this}\SetKwData{Up}{up} \SetKwFunction{Union}{Union}\SetKwInOut{Input}{Input}\SetKwInOut{Output}{Output}	
\Input{Image set \(V\), text set \(T\), constraint \(\epsilon\), norm \(\ell_p\).} 
\Output{Perturbation \(\delta\).} 
Initialize \(\delta \gets \boldsymbol {0}\)
\BlankLine 
% \emph{special treatment of the first line}\; 
\For{\(v\) in \(V\)}
{
getting \(Y = \{y_j\}_{j=1}^{n}\) and \(Y' = \{y'_{j}\}_{j=1}^{k}\)

\(v \gets v + \delta\)

\(r \gets \boldsymbol {0}\)

\While{\(I(E_v(v + (1+\eta)\cdot r),E_t(T),k)==1\)}
{
\(\hat{v} \gets v + r \)

\(y'_{min} \gets \underset{y'\in Y'}{argmin} f_{y'}(\hat{v})\)

\(y_{max} \gets \underset{y\in Y}{argmax} f_{y}(\hat{v})\)

\(r \gets r + \frac{\left(\nabla_r f_{y'_{min}}(\hat{v})-\nabla_r f_{y_{max}}(\hat{v})\right)\cdot  \left(f_{y_{max}}(\hat{v})-f_{y'_{min}}(\hat{v})\right) }{ \left \| \nabla_r f_{y'_{min}}(\hat{v})-\nabla_r f_{y_{max}}(\hat{v}) \right \|_2^2}\)

}

\(\delta \gets P_{\ell_p}(\delta+(1+\eta)\cdot r, \epsilon)\)

% \(\delta \gets clamp_{(0,1)}(\delta)\)

}
return \( \delta \)
\caption{Global TRA}
\label{alg:globalira} 
\end{algorithm}
\DecMargin{1em} 
%%%%%%%%%%%%%%%%%%%%%%%%%%%%%%%%%%%%%%%%%%%%%%%%%%%%%%%%%%%%%%%%%%%%%%%%
% TRA global版 结束
%%%%%%%%%%%%%%%%%%%%%%%%%%%%%%%%%%%%%%%%%%%%%%%%%%%%%%%%%%%%%%%%%%%%%%%%
% √

%%%%%%%%%%%%%%%%%%%%%%%%%%%%%%%%%%%%%%%%%%%%%%%%%%%%%%%%%%%%%%%%%%%%%%%%
% IRA global版 开始
%%%%%%%%%%%%%%%%%%%%%%%%%%%%%%%%%%%%%%%%%%%%%%%%%%%%%%%%%%%%%%%%%%%%%%%%
\IncMargin{1em}
\begin{algorithm}[b]
\SetKwData{Left}{left}\SetKwData{This}{this}\SetKwData{Up}{up} \SetKwFunction{Union}{Union}\SetKwInOut{Input}{Input}\SetKwInOut{Output}{Output}	
\Input{Image set \(V\), text set \(T\), constraint \(\epsilon\), norm \(\ell_p\).} 
\Output{Perturbation \(\delta\).} 
Initialize \(\delta \gets \boldsymbol {0}\)
\BlankLine 
% \emph{special treatment of the first line}\; 
\For{\(t\) in \(T\)}
{
getting \(y\) and \(Y' = \{y'_{j}\}_{j=1}^{k}\)

\(\tilde{V} = \{\tilde{v}_i\}_{i\in y\cup Y'} \gets \{v_i + \delta\}_{i\in y\cup Y'}\)

\(r \gets \boldsymbol {0}\)

\While{\(I(E_t(t),E_v(\tilde{V}+(1+\eta)\cdot r),k)==1\)}
{
\(\hat{V} =\{\hat{v}_i\}_{i\in y\cap Y'} \gets \tilde{V}+r\)

\(y'_{min} \gets \underset{y'\in Y'}{argmin}f_{y'}(t)\) \tcp{\(f_i(t)=E_v(\hat{v}_{i})\cdot E_t(t)\)}

\(r \gets r + \frac{\left(\nabla_r f_{y'_{min}}(t)-\nabla_r f_{y}(t)\right)\cdot  \left(f_{y}(t)-f_{y'_{min}}(t)\right) }{ \left \| \nabla_r f_{y'_{min}}(t)-\nabla_r f_{y}(t) \right \|_2^2}\)

}

\(\delta \gets P_{\ell_p}(\delta+(1+\eta)\cdot r, \epsilon)\)

% \(\delta \gets clamp_{(0,1)}(\delta)\)

}
return \( \delta \)
\caption{Global IRA}
\label{alg:globaltra} 
\end{algorithm}
\DecMargin{1em} 
%%%%%%%%%%%%%%%%%%%%%%%%%%%%%%%%%%%%%%%%%%%%%%%%%%%%%%%%%%%%%%%%%%%%%%%%
% IRA global版 结束
%%%%%%%%%%%%%%%%%%%%%%%%%%%%%%%%%%%%%%%%%%%%%%%%%%%%%%%%%%%%%%%%%%%%%%%%
% √

\section{Additional experiments based on CLIP} \label{Additional experiments based on CLIP}
% √

%%%%%%%%%%%%%%%%%%%%%%%%%%%%%%%%%%%%%%%%%%%%%%%%%%%%%%%%%%%%%%%%%%%%%%%%
% 表格对比Advclip与UAP IRA patch开始  图像分类
%%%%%%%%%%%%%%%%%%%%%%%%%%%%%%%%%%%%%%%%%%%%%%%%%%%%%%%%%%%%%%%%%%%%%%%%
\begin{table}[]
\caption{Comparing between AdvCLIP and UAP\(^{patch}_{IRA}\), based on the ImageNet, CIFAR-100 and CIFAR-10 datasets, the accuracy is reported. Lower is better.}
\label{tab:advclip2_}
\setlength{\tabcolsep}{3pt}
\renewcommand\arraystretch{1.2}
\begin{tabular}{ll|cc|cc|cc}
\toprule
\multirow{2}{*}{\parbox{1.2cm}{Visual \\Encoder}} & \multirow{2}{*}{Method} & \multicolumn{2}{c|}{ImageNet}    & \multicolumn{2}{c|}{CIFAR100}    & \multicolumn{2}{c}{CIFAR10}                    \\
\cmidrule(r){3-4} \cmidrule(r){5-6} \cmidrule(r){7-8}  
\multicolumn{2}{c|}{}     & Top1     & Top5    & Top1   &Top5   & Top1   & Top5  \\
\midrule
\multirow{3}{*}{ViT-B/16} & w/o atk & 62.42 & 86.74 & 66.56 & 88.49 & 90.10 & 99.07 \\
                        & AdvCLIP & 5.98  & 14.80 & 8.91  & 45.04 & 81.90 & 97.01 \\
                        & UAP\(^{patch}_{IRA}\)     & \textbf{0.13}  &\textbf{ 0.54}  & \textbf{0.94}  & \textbf{5.66}  & \textbf{9.88} & \textbf{47.03} \\
\midrule                   
\multirow{3}{*}{ViT-B/32} & w/o atk & 57.50 & 83.57 & 62.27 & 86.98 & 88.34 & 99.24 \\
                        & AdvCLIP & 7.98  & 27.40 & 31.92 & 51.84 & 64.45 & 96.51 \\
                        & UAP\(^{patch}_{IRA}\) & \textbf{2.45}  & \textbf{7.05}  & \textbf{8.90}  & \textbf{22.57} & \textbf{10.85} & \textbf{81.12} \\
\midrule
\multirow{3}{*}{ViT-L/14} & w/o atk & 69.74 & 90.15 & 75.72 & 93.06 & 95.19 & 99.52 \\
                        & AdvCLIP & 1.54  & 7.61  & 7.42  & 32.30 & 73.87 & 97.95 \\
                        & UAP\(^{patch}_{IRA}\) & \textbf{0.21}  & \textbf{0.74 } & \textbf{4.52}  & \textbf{14.96} &\textbf{ 22.71 }& \textbf{73.67} \\              
\bottomrule
\end{tabular}
\end{table}
%%%%%%%%%%%%%%%%%%%%%%%%%%%%%%%%%%%%%%%%%%%%%%%%%%%%%%%%%%%%%%%%%%%%%%%%
% 表格对比Advclip与UAP IRA patch结束  图像分类
%%%%%%%%%%%%%%%%%%%%%%%%%%%%%%%%%%%%%%%%%%%%%%%%%%%%%%%%%%%%%%%%%%%%%%%%
% √

%%%%%%%%%%%%%%%%%%%%%%%%%%%%%%%%%%%%%%%%%%%%%%%%%%%%%%%%%%%%%%%%%%%%%%%%
% 表格对比Advclip与UAP IRA patch开始  迁移
%%%%%%%%%%%%%%%%%%%%%%%%%%%%%%%%%%%%%%%%%%%%%%%%%%%%%%%%%%%%%%%%%%%%%%%%
\begin{table*}[]
\caption{Comparing between AdvCLIP and UAP\(^{patch}_{IRA}\), based on the NUS-WIDE, Pascal-Sentence and Wikipedia datasets.}

\label{tab:advclip3_}
\renewcommand\arraystretch{1.1}
\begin{tabular}{ll|ccccccccc}
\toprule
\multirow{2}{*}{\parbox{1.2cm}{Visual \\Encoder}} & \multirow{2}{*}{Metric} & \multicolumn{3}{c}{NUS-WIDE} & \multicolumn{3}{c}{Pascal-Sentence} & \multicolumn{3}{c}{Wikipedia} \\
\cmidrule(r){3-5} \cmidrule(r){6-8} \cmidrule(r){9-11}  
                         &                         & w/o atk  & AdvCLIP  & UAP\(^{patch}_{IRA}\)   & w/o atk  & AdvCLIP & UAP\(^{patch}_{IRA}\)  & w/o atk   & AdvCLIP  & UAP\(^{patch}_{IRA}\)   \\
\midrule
\multirow{5}{*}{ViT-B/16}  & TR@1                    & 66.90    & 36.75    & \textbf{12.65}  & 67.00    & 7.00    & \textbf{6.50}  & 61.69     & 14.72    & \textbf{9.96}   \\
                         & IR@1                    & 61.80    & 51.25    & \textbf{38.15}  & 68.50    & 10.00   & \textbf{8.50}  & 66.67     & 42.64    & \textbf{32.68}  \\
                         & AVG                     & 64.35    & 44.00    & \textbf{25.4}   & 67.75    & 8.50    & \textbf{7.50}  & 64.18     & 28.68    & \textbf{21.32}  \\
                         & Top1                    & 77.85    & 49.65    & \textbf{12.15}  & 77.00    & \textbf{5.50}    & 8.00  & 64.01     & 27.95    & \textbf{8.41}   \\
                         & Top5                    & 98.75    & 72.95    & \textbf{55.75}  & 96.00    & 39.50   & \textbf{32.00} & 94.35     & 73.15    & \textbf{55.60}  \\
\midrule
\multirow{5}{*}{ViT-B/32}  & TR@1                    & 69.25    & 34.15    & \textbf{31.85}  & 72.00    & 35.00   & \textbf{14.50} & 58.23     & 31.60    & \textbf{19.91}  \\
                         & IR@1                    & 60.25    & \textbf{47.60}    & 49.95  & 70.50    & 47.00   & \textbf{15.00} & 62.77     & 48.70    & \textbf{42.64}  \\
                         & AVG                     & 64.75    & \textbf{40.88}    & 40.90  & 71.25    & 41.00   & \textbf{14.75} & 60.5      & 40.15    & \textbf{31.28}  \\
                         & Top1                    & 77.65    & 32.60    & \textbf{23.90}  & 72.00    & 52.00   & \textbf{10.00} & 62.09     & 35.33    & \textbf{21.99}  \\
                         & Top5                    & 98.55    & 87.05    & \textbf{65.75}  & 94.50    & 88.00   & \textbf{48.50} & 93.91     & 81.30    & \textbf{62.61}  \\
\midrule
\multirow{5}{*}{ViT-L/14}  & TR@1                    & 67.50    & 37.15    & \textbf{12.10}  & 65.50    & 17.00   & \textbf{1.50}  & 62.55     & 40.48    & \textbf{10.17}  \\
                         & IR@1                    & 61.15    & 50.10    & \textbf{18.40}  & 68.00    & 9.50    & \textbf{5.50}  & 62.77     & 50.43    & \textbf{26.41}  \\
                         & AVG                     & 64.325   & 43.625   & \textbf{15.25}  & 66.75    & 13.25   & \textbf{3.50}  & 62.66     & 45.455   & \textbf{18.29}  \\
                         & Top1                    & 76.35    & 16.15    & \textbf{11.15}  & 75.00    & 22.00   & \textbf{5.00}  & 67.09     & 35.36    & \textbf{8.98}   \\
                         & Top5                    & 98.20    & 76.05    & \textbf{51.00}  & 94.20    & 69.50   & \textbf{26.00} & 94.23     & 74.63    & \textbf{42.31}  \\
\bottomrule
\end{tabular}
\end{table*}
%%%%%%%%%%%%%%%%%%%%%%%%%%%%%%%%%%%%%%%%%%%%%%%%%%%%%%%%%%%%%%%%%%%%%%%%
% 表格对比Advclip与UAP IRA patch结束  迁移
%%%%%%%%%%%%%%%%%%%%%%%%%%%%%%%%%%%%%%%%%%%%%%%%%%%%%%%%%%%%%%%%%%%%%%%%
% √

In the main text, UAP\(^{patch}_{TIRA}\) and AdvCLIP \cite{advclip} have been thoroughly compared.
% √
In fact, using IRA alone can also achieve stable attack effects.
% √
Therefore, we compare UAP\(^{patch}_{IRA}\) with AdvCLIP, keeping all other experimental settings unchanged.
% √
The experimental results are shown in Tab~\ref{tab:advclip1_}, Tab~\ref{tab:advclip2_}, and Tab~\ref{tab:advclip3_}. 
% √
Except for a few cases in Tab~\ref{tab:advclip3_} where UAP\(^{patch}_{IRA}\) is slightly inferior to AdvCLIP, in most situations, UAP\(^{patch}_{IRA}\) achieves more effective and stable attack effects.
% √

\section{Generating Global Universal Adversarial Perturbation} 
\label{Generating Global Universal Adversarial Perturbation}
% √

This section explains how to generate universal global adversarial perturbations and provides the visualizations.
% √

\subsection{Noise Constraints}
Global adversarial attacks can modify any pixel in an image (equivalently understood as adding noise to the image) \cite{fgsm,pgd}. 
% √
Still, modifying pixels (the magnitude of noise) is constrained to ensure it is difficult for human detection. 
% √
And it does not alter the original semantic information of the image.
% √
Considering the average \(\ell_2\) and \(\ell_\infty\) norms of images in the ImageNet validation set are approximately \(5 \times 10^4\) and \(250\), we set \(\epsilon=2000\) for \(\ell_2\) norm and \(\epsilon=10\) for \(\ell_\infty\) norm to obtain perturbations whose norm is significantly smaller than the original image norm \cite{deepfool}. 
% √
The valid range for pixel values here is \([0,255]\), without considering normalization.
% √
\(\epsilon\) represents the noise constraint for global perturbations, i.e., \(\left \|  \delta \right \| _p \le \epsilon\).
% √
For any given noise \(\delta\), to ensure \(\left \|  \delta \right \| _p \le \epsilon\), a projection operation should be performed \cite{pgd}.
% √
The projection strategy based on the \(\ell_2\) norm is:
% √
\begin{equation}
P_{\ell_2}(\delta, \epsilon)=\begin{cases}
  \epsilon\cdot\frac{\delta}{\left \| \delta \right \|_2 } & \text{ if } \left \| \delta \right \|_2>\epsilon \\
 \delta& \text{ if } \left \| \delta \right \|_2\le\epsilon
\end{cases}
\end{equation}
% √
where \(P_{\ell_2}\) ensures that if \(\left \| \delta \right \|_2 > \epsilon\), i.e., the noise exceeds the valid boundary, it will be projected back into the defined noise range.
% √
The projection strategy based on the \(\ell_\infty\) norm is:
% √
\begin{equation}
P_{\ell_\infty}(\delta, \epsilon)=clamp_{(-\epsilon,\epsilon)}(\delta)
\end{equation}
% √
where \(clamp_{(-\epsilon,\epsilon)}\) represents limiting the modification range of each pixel by clipping each element value in \(\delta\) to be between \(-\epsilon\) and \(\epsilon\).
% √

\subsection{Global Universal Attack Algorithms} 
% √

%%%%%%%%%%%%%%%%%%%%%%%%%%%%%%%%%%%%%%%%%%%%%%%%%%%%%%%%%%%%%%%%%%%%%%%%
% 6种全局扰动可视化 开始
%%%%%%%%%%%%%%%%%%%%%%%%%%%%%%%%%%%%%%%%%%%%%%%%%%%%%%%%%%%%%%%%%%%%%%%%
\begin{figure}[]
\centering
\includegraphics[width=\linewidth]{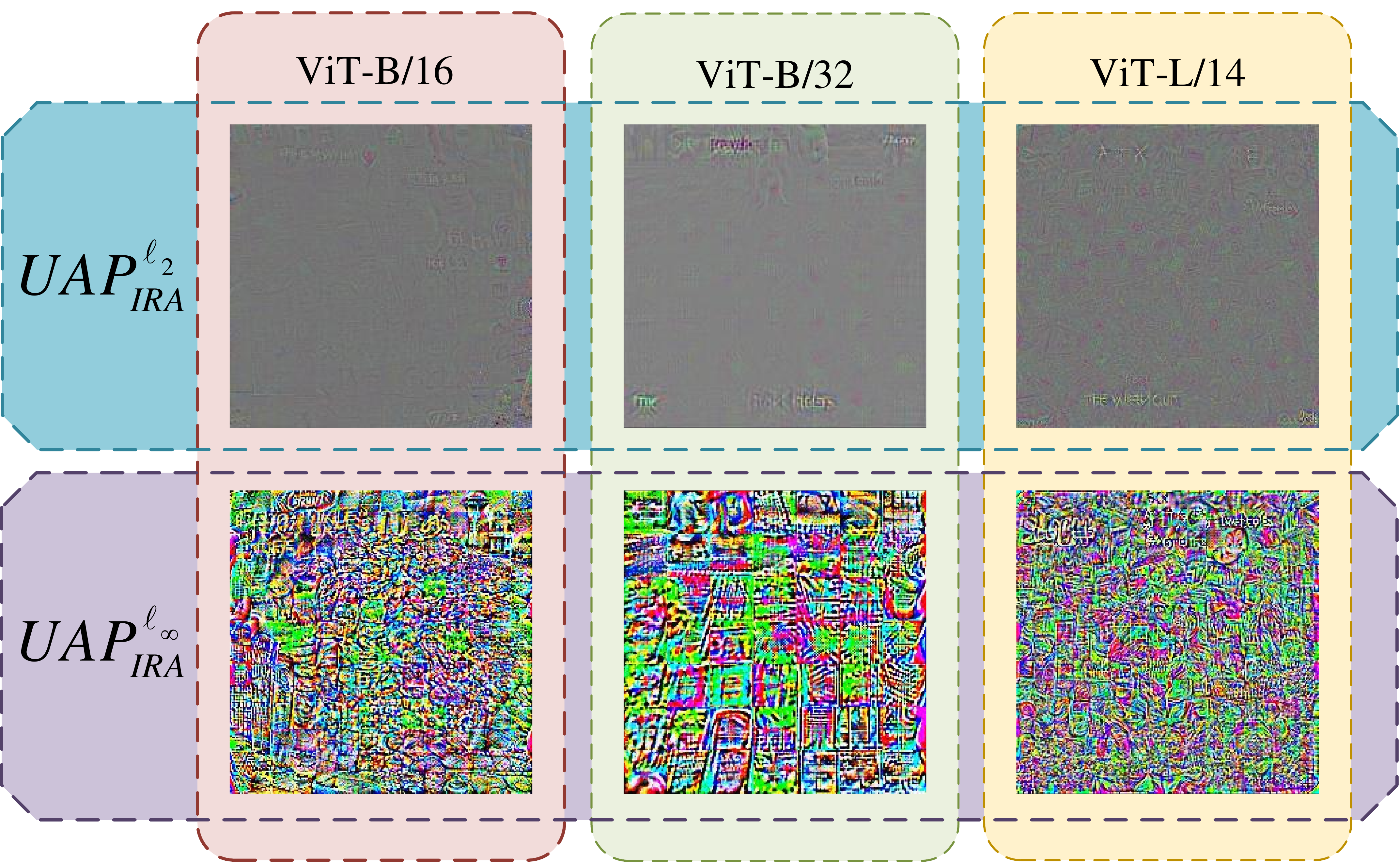}
\caption{Visualization of six universal adversarial global perturbations.}
\label{fig:global-per}
\Description{Global universal adversarial perturbations. }
\end{figure}
%%%%%%%%%%%%%%%%%%%%%%%%%%%%%%%%%%%%%%%%%%%%%%%%%%%%%%%%%%%%%%%%%%%%%%%%
% 6种全局扰动可视化 结束
%%%%%%%%%%%%%%%%%%%%%%%%%%%%%%%%%%%%%%%%%%%%%%%%%%%%%%%%%%%%%%%%%%%%%%%%
% √

%%%%%%%%%%%%%%%%%%%%%%%%%%%%%%%%%%%%%%%%%%%%%%%%%%%%%%%%%%%%%%%%%%%%%%%%
% 6种全局扰动 开始
%%%%%%%%%%%%%%%%%%%%%%%%%%%%%%%%%%%%%%%%%%%%%%%%%%%%%%%%%%%%%%%%%%%%%%%%
\begin{table*}[]
\caption{The effects of six universal adversarial global perturbations.}
\label{tab:globalattack}
\renewcommand\arraystretch{1.1}
\begin{tabular}{ll|ccc|ccc|ccc|ccc}
\toprule
\multirow{3}{*}{\parbox{1.2cm}{Visual \\Encoder}} & \multirow{3}{*}{Method} & \multicolumn{6}{c|}{Flickr30k(1K test set)}                         & \multicolumn{6}{c}{MS COCO(5K test set)}                        \\
\multicolumn{2}{c|}{}                        & \multicolumn{3}{c}{TR} & \multicolumn{3}{c|}{IR} & \multicolumn{3}{c}{TR} & \multicolumn{3}{c}{IR} \\
\cmidrule(r){3-5} \cmidrule(r){6-8} \cmidrule(r){9-11} \cmidrule(r){12-14} 
\multicolumn{2}{c|}{}                        & R@1     & R@5    & R@10   & R@1     & R@5    & R@10   & R@1     & R@5    & R@10   & R@1     & R@5    & R@10   \\
\midrule
\multirow{3}{*}{ViT-B/16} & w/o atk & 82.20 & 96.70 & 98.90 & 62.12 & 85.72 & 91.94 & 52.40 & 76.78 & 84.66 & 33.07 & 58.35 & 69.03   \\
                          & UAP\(^{\ell_\infty}_{IRA}\) & 38.80 & 62.30    & 71.80  & 9.40 & 17.31 & 22.56 & 15.76  & 31.80  & 40.00 & 6.73 & 14.39 & 18.78   \\
                          & UAP\(^{\ell_2}_{IRA}\)& \textbf{36.00}    & \textbf{57.70}  & \textbf{67.4}  & \textbf{8.36}  & \textbf{15.64}  & \textbf{21.04}  & \textbf{14.30}   & \textbf{27.72}  & \textbf{34.78} & \textbf{5.45}  & \textbf{11.62}  & \textbf{15.80}  \\
\midrule
\multirow{3}{*}{ViT-B/32} & w/o atk & 78.60 & 94.90 & 98.30 & 58.80 & 83.56 & 90.00 & 50.18 & 75.02 & 83.50 & 30.48 & 55.99 & 66.87   \\
                          & UAP\(^{\ell_\infty}_{IRA}\) & 41.00 & 64.00 & 70.70 & 13.22  & 24.46 & 30.86 & 17.70 & 32.60 & 40.96 & 7.59 & 16.70  & 22.35 \\
                          &UAP\(^{\ell_2}_{IRA}\) & \textbf{13.70}  & \textbf{25.10}  & \textbf{32.20}  & \textbf{12.00}  & \textbf{21.12}   & \textbf{26.04}  & \textbf{5.80}     & \textbf{12.26}   & \textbf{16.40} & \textbf{6.06}   & \textbf{12.92}  & \textbf{17.10}    \\
\midrule
\multirow{3}{*}{ViT-L/14} & w/o atk & 85.30 & 97.40 & 99.20 & 64.84 & 87.22 & 92.04 & 56.34 & 79.40 & 86.58 & 36.51 & 61.08 & 71.17   \\
                          & UAP\(^{\ell_\infty}_{IRA}\) & 42.40 & 64.10 & 69.30  & \textbf{8.20} & \textbf{14.44} & \textbf{18.60} & 16.32  & 31.34  & 38.48 & \textbf{4.43} & \textbf{8.55} & \textbf{10.78}  \\
                          & UAP\(^{\ell_2}_{IRA}\) & \textbf{24.90}  & \textbf{39.80}  & \textbf{46.70}  & 14.12   & 24.22   & 29.16  & \textbf{7.76}  & \textbf{16.16}   & \textbf{20.74} & 5.19 & 10.52  & 13.37 \\
\bottomrule
\end{tabular}
\end{table*}
%%%%%%%%%%%%%%%%%%%%%%%%%%%%%%%%%%%%%%%%%%%%%%%%%%%%%%%%%%%%%%%%%%%%%%%%
% 6种全局扰动 结束
%%%%%%%%%%%%%%%%%%%%%%%%%%%%%%%%%%%%%%%%%%%%%%%%%%%%%%%%%%%%%%%%%%%%%%%%
% √

We revise the pseudocode for the IRA and TRA to generate universal adversarial global perturbations, as shown in Alg~\ref{alg:globalira} and Alg~\ref{alg:globaltra}, with three main modifications: 
% √
(1) When adding the perturbation \(\delta\) or \(r\) to an image, the masking matrix \(m\) is unnecessary which is used to determine the location of the patch. 
% √
Instead, the image and the perturbation are directly added together. 
% √
(2) 
The adversarial patch directly replaces pixel values in a specified area of the image, it only needs to ensure that the pixel values of the patch block are within the valid range \cite{lavan}. 
% √
The global perturbation modifies all pixel values of the original image, i.e. the noise is superimposed on the original image, thus necessitating a projection function \(P_{\ell_p}\) used to limit the extent of modifications.
% √
(3) During the testing phase (i.e., after the optimization of the universal perturbation is complete and it enters the usage stage), directly adding the universal global perturbation to each image cannot guarantee that the pixel values of each adversarial image are within the valid range. 
% √
% Therefore, we must manually clip each of them.
Due to this, the universal global perturbation added to each clean image undergoes varying degrees of clipping. 
This variability in modification might partially account for the lesser effectiveness of universal global perturbations compared to universal adversarial patches.
% √

\subsection{Visualization}

We use 1000 images and 5000 texts from the Flickr30k \cite{flickr30k} dataset as surrogate data to generate six universal global adversarial perturbations based on three CLIP \cite{clip} pre-trained models and two norm constraints. 
% √
We provide the results of the attack effectiveness, as seen in Tab~\ref{tab:globalattack}, along with visualization of the perturbations in Fig~\ref{fig:global-per}.
% √

\section{Additional experiments based on BEiT3} 
\label{Additional experiments based on BEiT3}
% √
We further conduct detailed experiments on BEiT3 \cite{beit3}. 
% √
First, some configurations need to be determined:
% √
(1)
We consistently use TIRA, in which TRA and IRA are set with \(k=10\) because the Text Retrieval R@10 and Image Retrieval R@10 metrics will be evaluated.
% √
(2)
For generating universal adversarial patches, we set the patch size to \(66\times 66\) (BEiT3 requires an input image size of \(384\times 384\), thus the patch covers 3\% of the image area).
% √
(3)
The patch is located at the bottom right corner of the image, with the bottom right corner of the patch having an offset of \((x,y)=(-24,-24)\) relative to the bottom right corner of the image.
% √
(5)
For generating universal global perturbations, we use \(\ell_2\) norm to constraint the noise, setting \(\epsilon = 2000\).
% √

The ``BEiT3-base'' model is a pre-trained model obtained through the Masked Data Modeling (MDM) pre-training task. 
% √
It is further fine-tuned on the Image-Text Contrastive (ITC) task to obtain the ``BEiT3-base-itc'' pre-trained model.
% √
Building on ``BEiT3-base-itc'', the model is fine-tuned separately on the MS COCO dataset \cite{coco} and the Flickr30k dataset \cite{flickr30k}, resulting in two models specifically for image-text retrieval task, which we denote as \textbf{\(M1\)} and \textbf{\(M2\)}, respectively.
We consistently conduct the white-box attack to generate universal perturbation based on the \(M1\). 
% √
\(M2\) is reserved for testing the cross-model transferability of the universal perturbation.

\begin{table}[t]
\caption{The effects of universal global perturbations on BEiT3.}
\label{tab:globalbeit3}
\setlength{\tabcolsep}{3.8pt}
\begin{tabular}{lll|ccc|ccc}
\toprule
\multirow{2}{*}{MD} & \multirow{2}{*}{DS} & \multirow{2}{*}{Patch}  & \multicolumn{3}{c}{Text Retrieval}  & \multicolumn{3}{c}{Image Retrieval}    \\
\cmidrule(r){4-6} \cmidrule(r){7-9} 
\multicolumn{3}{c|}{}                        & R@1     & R@5    & R@10   & R@1     & R@5    & R@10   \\
\midrule
\multirow{6}{*}{\(M1\)}    & \multirow{3}{*}{\(D1\)}      & w/o atk     & 78.98 & 94.38 & 97.20  & 61.36 & 84.61 & 90.74 \\
                           &                            & \(Noise1\)  & 17.72 & 29.90 & 35.48 & 21.85 & 40.62 & 50.00  \\
                           &                            & \(Noise2\)  & 18.94 & 30.96 & 36.68 & 22.92 & 42.15 & 51.55 \\
\cmidrule(r){2-9}
                           & \multirow{3}{*}{\(D2\)} & w/o atk   & 93.60 & 99.30 & 99.80  & 82.90 & 96.54 & 98.46 \\
                           &                            & \(Noise1\)  & 44.20 & 63.10 & 69.20 & 43.00 & 67.10 & 76.10\\
                           &                            & \(Noise2\)  & 40.80 & 56.30 & 63.80 & 44.66 & 68.50 & 76.56 \\
\midrule
\multirow{6}{*}{\(M2\)}    & \multirow{3}{*}{\(D1\)}      & w/o atk     & 71.00 & 90.16 & 94.44  & 52.79 & 77.12 & 85.07 \\
                           &                             &\(Noise1\)   & 22.34 & 36.86 & 43.92 & 20.30 & 39.17 & 49.20 \\
                           &                             &\(Noise2\)  & 22.62 & 37.36 & 45.00 & 19.70 & 37.29 & 46.71 \\
\cmidrule(r){2-9}
                           & \multirow{3}{*}{\(D2\)} & w/o atk     & 96.30 & 99.70 & 100.00 & 86.14 & 97.68 & 98.82 \\
                           &                            & \(Noise1\)  & 56.60 & 75.10 & 79.60 & 57.58 & 82.08 & 87.84 \\
                           &                            & \(Noise2\)  & 53.10 & 69.70 & 76.00 & 56.24 & 78.72 & 85.90 \\
\bottomrule
\end{tabular}
\end{table}

\subsection{Generating Universal Adversarial Patch}

We use 1000 training images from the MS COCO dataset and their matching texts, as well as 1000 training images and their matching texts from the Flickr30k dataset, to generate two universal adversarial patches, respectively denoted as \(Patch1\) and \(Patch2\). 
% √
A comprehensive evaluation of their attack effectiveness is presented in Tab~\ref{tab:patchbeit3}.
% √
The first column indicates the model under test, and the second row specifies the dataset under test, where \(D1\) represents the MS COCO test set, and \(D2\) represents the Flickr30k test set. 
% √
The third row indicates which patch is being applied to the entire set of test images.
% √
The experimental results validate that our universal adversarial patches possess stable and potent attack effectiveness, as well as good cross-dataset and cross-model transferability.
% √

\subsection{Generating Universal Global Perturbation}
% √

We use 1000 training images from the MS COCO dataset and their matching texts, as well as 1000 training images and their matching texts from the Flickr30k dataset, to generate two universal adversarial global perturbations, respectively denoted as \(Noise1\) and \(Noise2\). 
% √
A comprehensive evaluation of their attack effectiveness is presented in Tab~\ref{tab:globalbeit3}.
% √
The experimental results validate that our universal adversarial global perturbations possess stable and potent attack effectiveness, as well as good cross-dataset and cross-model transferability.
% √

\printbibliography

@String{Computing = "Computing" }

@String{Computer = "{IEEE} Computer" }

@INPROCEEDINGS{deepfool,
  author={Moosavi-Dezfooli, Seyed-Mohsen and Fawzi, Alhussein and Frossard, Pascal},
  booktitle={Proceedings of the IEEE Conference on Computer Vision and Pattern Recognition (CVPR)},
  pages={2574--2582},
  title={{DeepFool}: A Simple and Accurate Method to Fool Deep Neural Networks}, 
  year={2016},
}

@INPROCEEDINGS {mrl,
author = {Yuan, Xin and Lin, Zhe and Kuen, Jason and Zhang, Jianming and Wang, Yilin and Maire, Michael and Kale, Ajinkya and Faieta, Baldo},
booktitle={Proceedings of the IEEE/CVF Conference on Computer Vision and Pattern Recognition (CVPR)},
pages={6995--7004},
title = {Multimodal Contrastive Training for Visual Representation Learning},
year = {2021},
}

@article{coop,
author = {Zhou, Kaiyang and Yang, Jingkang and Loy, Chen Change and Liu, Ziwei},
title = {Learning to Prompt for Vision-Language Models},
year = {2022},
pages={2337--2348},
journal = {International Journal of Computer Vision (IJCV)},
}

@INPROCEEDINGS{cocoop,
  author={Zhou, Kaiyang and Yang, Jingkang and Loy, Chen Change and Liu, Ziwei},
  booktitle={CVPR}, 
  title={Conditional Prompt Learning for Vision-Language Models}, 
  booktitle={Proceedings of the IEEE/CVF Conference on Computer Vision and Pattern Recognition (CVPR)},
  pages={16816--16825},
  year={2022},}

@InProceedings{vlbackdoor,
author={Jiawei, Liang and Siyuan, Liang and Man, Luo and Aishan, Liu and Dongchen, Han and Ee-Chien, Chang and Xiaochun, Cao},
title={VL-Trojan: Multimodal Instruction Backdoor Attacks against Autoregressive Visual Language Models},
booktitle={arXiv preprint arXiv:2402.13851},
year={2024},
}

@inproceedings{uniter,
  title={Uniter: Universal image-text representation learning},
  author={Chen, Yen-Chun and Li, Linjie and Yu, Licheng and El Kholy, Ahmed and Ahmed, Faisal and Gan, Zhe and Cheng, Yu and Liu, Jingjing},
  booktitle={Proceedings of the 16th European Conference on Computer Vision (ECCV)},
  pages={104--120},
  year={2020},
}

@inproceedings{oscar,
  title={Oscar: Object-semantics aligned pre-training for vision-language tasks},
  author={Li, Xiujun and Yin, Xi and Li, Chunyuan and Zhang, Pengchuan and Hu, Xiaowei and Zhang, Lei and Wang, Lijuan and Hu, Houdong and Dong, Li and Wei, Furu and others},
  booktitle={Proceedings of the 16th European Conference on Computer Vision (ECCV)},
pages={121--137},
  year={2020},
}

@inproceedings{vlmixer,
  title={Vlmixer: Unpaired vision-language pre-training via cross-modal cutmix},
  author={Wang, Teng and Jiang, Wenhao and Lu, Zhichao and Zheng, Feng and Cheng, Ran and Yin, Chengguo and Luo, Ping},
booktitle = 	 {Proceedings of the International Conference on Machine
 Learning (ICML)},
  pages={22680--22690},
  year={2022},
}

@inproceedings{vinvl,
  title={Vinvl: Revisiting visual representations in vision-language models},
  author={Zhang, Pengchuan and Li, Xiujun and Hu, Xiaowei and Yang, Jianwei and Zhang, Lei and Wang, Lijuan and Choi, Yejin and Gao, Jianfeng},
  booktitle={Proceedings of the IEEE/CVF Conference on Computer Vision and Pattern Recognition (CVPR)},
  pages={5579--5588},
  year={2021}
}

@inproceedings{bert,
  title={Bert: Pre-training of deep bidirectional transformers for language understanding},
  author={Devlin, Jacob and Chang, Ming-Wei and Lee, Kenton and Toutanova, Kristina},
  booktitle={arXiv preprint arXiv:1810.04805},
  year={2018}
}

@inproceedings{uap,
  title={Universal adversarial perturbations},
  author={Moosavi-Dezfooli, Seyed-Mohsen and Fawzi, Alhussein and Fawzi, Omar and Frossard, Pascal},
  booktitle={Proceedings of the IEEE Conference on Computer Vision and Pattern Recognition (CVPR)},
  pages={1765--1773},
  year={2017}
}

@inproceedings{sguap,
  title={Universal adversarial training},
  author={Shafahi, Ali and Najibi, Mahyar and Xu, Zheng and Dickerson, John and Davis, Larry S and Goldstein, Tom},
booktitle={Proceedings of the Association for the Advancement of Artificial Intelligence (AAAI)},
  pages={5636--5643},
  year={2020}
}

@inproceedings{gap,
  title={Generative adversarial perturbations},
  author={Poursaeed, Omid and Katsman, Isay and Gao, Bicheng and Belongie, Serge},
  booktitle={Proceedings of the IEEE Conference on Computer Vision and Pattern Recognition (CVPR)},
  pages={742--751},
  year={2018}
}

@inproceedings{fff,
  title={Fast feature fool: A data independent approach to universal adversarial perturbations},
  author={Mopuri, Konda Reddy and Garg, Utsav and Babu, R Venkatesh},
  booktitle={arXiv preprint arXiv:1707.05572},
  year={2017}
}

@inproceedings{dfuap,
  title={Ask, acquire, and attack: Data-free uap generation using class impressions},
  author={Mopuri, Konda Reddy and Uppala, Phani Krishna and Babu, R Venkatesh},
  booktitle={Proceedings of the 15th European Conference on Computer Vision (ECCV)},
  pages={19--34},
  year={2018}
}

@article{tsne,
  title={Visualizing data using t-SNE},
  author={Van der Maaten, Laurens and Hinton, Geoffrey},
  journal={Journal of Machine Learning Research (JMLR)},
  year={2008}
}

@inproceedings{wikipedia,
  title={Collecting image annotations using amazon’s mechanical turk},
  author={Rashtchian, Cyrus and Young, Peter and Hodosh, Micah and Hockenmaier, Julia},
booktitle={Proceedings of the NAACL HLT 2010 workshop on creating speech and language data with Amazon’s Mechanical Turk (NAACL-HLTW)},
  pages={139--147},
  year={2010}
}

@inproceedings{pascal-sentence,
  title={A new approach to cross-modal multimedia retrieval},
  author={Rasiwasia, Nikhil and Costa Pereira, Jose and Coviello, Emanuele and Doyle, Gabriel and Lanckriet, Gert RG and Levy, Roger and Vasconcelos, Nuno},
 booktitle={Proceedings of the 18th ACM international conference on Multimedia (ACM MM)},
  pages={251--260},
  year={2010}
}

@inproceedings{nus-wide,
  title={Nus-wide: a real-world web image database from national university of singapore},
  author={Chua, Tat-Seng and Tang, Jinhui and Hong, Richang and Li, Haojie and Luo, Zhiping and Zheng, Yantao},
  booktitle={Proceedings of the ACM International Conference on Image and Video Retrieval (CIVR)},
  pages={1--9},  
  year={2009}
}

@inproceedings{li1,
author = {Li, Wenrui and Ma, Zhengyu and Deng, Liang-Jian and Wang, Penghong and Shi, Jinqiao and Fan, Xiaopeng},
title = {Reservoir Computing Transformer for Image-Text Retrieval},
year = {2023}, 
booktitle={Proceedings of the 31st ACM International Conference on Multimedia (ACM MM)},
pages={5605--5613}
}

@inproceedings{vit,
  title={An Image is Worth 16x16 Words: Transformers for Image Recognition at Scale},
  author={Dosovitskiy, Alexey and Beyer, Lucas and Kolesnikov, Alexander and Weissenborn, Dirk and Zhai, Xiaohua and Unterthiner, Thomas and Dehghani, Mostafa and Minderer, Matthias and Heigold, Georg and Gelly, Sylvain and others},
  booktitle={International Conference on Learning Representations (ICLR)},
  year = {2023}
}

@inproceedings{albef,
      title={Align before Fuse: Vision and Language Representation Learning with Momentum Distillation}, 
      author={Li, Junnan and Selvaraju, Ramprasaath and Gotmare, Akhilesh and Joty, Shafiq and Xiong, Caiming and Hoi, Steven Chu Hong},
      booktitle={Proceedings of the 35th
 International Conference on Neural Information Processing Systems (NeurIPS)},
pages={9694--9705},
year = {2021}
}

@InProceedings{tcl,
    author    = {Yang, Jinyu and Duan, Jiali and Tran, Son and Xu, Yi and Chanda, Sampath and Chen, Liqun and Zeng, Belinda and Chilimbi, Trishul and Huang, Junzhou},
    title     = {Vision-Language Pre-Training With Triple Contrastive Learning},
    booktitle = {Proceedings of the IEEE/CVF Conference on Computer Vision and Pattern Recognition (CVPR)},
pages={15671--15680},
year = {2022}
}

@inproceedings{beit3,
title={Image as a Foreign Language: {BEiT} Pretraining for Vision and Vision-Language Tasks},
author={Wang, Wenhui and Bao, Hangbo and Dong, Li and Bjorck, Johan and Peng, Zhiliang and Liu, Qiang and Aggarwal, Kriti and Mohammed, Owais Khan and Singhal, Saksham and Som, Subhojit and others},
booktitle={Proceedings of the IEEE/CVF Conference on Computer Vision and Pattern Recognition (CVPR)},
pages={19175--19186},
year={2023}
}

@inproceedings{beit2,
title={{BEiT v2}: Masked Image Modeling with Vector-Quantized Visual Tokenizers},
author={Zhiliang Peng and Li Dong and Hangbo Bao and Qixiang Ye and Furu Wei},
booktitle={arXiv preprint arXiv:2208.06366},
year = {2022}
}

@inproceedings{beit,
title={{BEiT}: {BERT} Pre-Training of Image Transformers},
author={Bao, Hangbo and Dong, Li and Piao, Songhao and Wei, Furu},
booktitle={International Conference on Learning Representations (ICLR)},
year={2022},
}

@InProceedings{vilt,
  title = 	 {ViLT: Vision-and-Language Transformer Without Convolution or Region Supervision},
  author =       {Kim, Wonjae and Son, Bokyung and Kim, Ildoo},
  booktitle = 	 {Proceedings of the International Conference on Machine
 Learning (ICML)},
pages={5583--5594},
  year = 	 {2021},
}

@inproceedings{vlmo,
 author = {Bao, Hangbo and Wang, Wenhui and Dong, Li and Liu, Qiang and Mohammed, Owais Khan and Aggarwal, Kriti and Som, Subhojit  and Piao, Songhao and Wei, Furu},
 booktitle = {Proceedings of the 36th
 International Conference on Neural Information Processing Systems (NeurIPS)},
 year = {2022},
pages={32897--32912},
title = {VLMo: Unified Vision-Language Pre-Training with Mixture-of-Modality-Experts},
}

@InProceedings{clip,
  title = 	 {Learning Transferable Visual Models From Natural Language Supervision},
  author =       {Radford, Alec and Kim, Jong Wook and Hallacy, Chris and Ramesh, Aditya and Goh, Gabriel and Agarwal, Sandhini and Sastry, Girish and Askell, Amanda and Mishkin, Pamela and Clark, Jack and Krueger, Gretchen and Sutskever, Ilya},
  booktitle = 	 {Proceedings of the International Conference on Machine
 Learning (ICML)},
pages={8748--8763},
  year = 	 {2021},
}

@inproceedings{fgsm,
title	= {Explaining and Harnessing Adversarial Examples},
author	= {Goodfellow, Ian J and Shlens, Jonathon and Szegedy, Christian},
year	= {2015},
booktitle	= {International Conference on Learning Representations (ICLR)},
}

@inproceedings{pgd,
title={Towards Deep Learning Models Resistant to Adversarial Attacks},
author={Madry, Aleksander and Makelov, Aleksandar and Schmidt, Ludwig and Tsipras, Dimitris and Vladu, Adrian},
booktitle={International Conference on Learning Representations (ICLR)},
year={2018},
}

@inproceedings{coattack,
author = {Zhang, Jiaming and Yi, Qi and Sang, Jitao},
title = {Towards Adversarial Attack on Vision-Language Pre-training Models},
year = {2022},
  booktitle={Proceedings of the 30th ACM International Conference on Multimedia (ACM MM)},
  pages={5005--5013},
}

@inproceedings{advclip,
author = {Zhou, Ziqi and Hu, Shengshan and Li, Minghui and Zhang, Hangtao and Zhang, Yechao and Jin, Hai},
title = {Adv{CLIP}: Downstream-agnostic Adversarial Examples in Multimodal Contrastive Learning},
  booktitle={Proceedings of the 31st ACM International Conference on Multimedia (ACM MM)},
  pages={6311--6320},
year = {2023},
}

@InProceedings{sga,
    author    = {Lu, Dong and Wang, Zhiqiang and Wang, Teng and Guan, Weili and Gao, Hongchang and Zheng, Feng},
    title     = {Set-level Guidance Attack: Boosting Adversarial Transferability of Vision-Language Pre-training Models},
  booktitle={Proceedings of the IEEE/CVF International Conference on Computer Vision (ICCV)},
  pages={102--111},
    year      = {2023},
}

@inproceedings{blip,
      title={BLIP: Bootstrapping Language-Image Pre-training for Unified Vision-Language Understanding and Generation}, 
      author={Li, Junnan and Li, Dongxu and Xiong, Caiming and Hoi, Steven},
      year={2022},
pages={12888--12900},
  booktitle = 	 {Proceedings of the International Conference on Machine
 Learning (ICML)},
}

@InProceedings{lavan,
  title = 	 {La{VAN}: Localized and Visible Adversarial Noise},
  author =       {Karmon, Danny and Zoran, Daniel and Goldberg, Yoav},
  booktitle = 	 {Proceedings of the International Conference on Machine
 Learning (ICML)},
  year = 	 {2018},
pages={2507--2515},
}

@article{cifar10,
  title={Learning multiple layers of features from tiny images},
  author={ Krizhevsky, A.  and  Hinton, G. },
  journal={Handbook of Systemic Autoimmune Diseases},
  year={2009},
}

@INPROCEEDINGS{imagenet,
  author={Deng, Jia and Dong, Wei and Socher, Richard and Li, Li-Jia and Kai, Li and Li, Fei-Fei},
    booktitle={Proceedings of the IEEE Conference on Computer Vision and Pattern Recognition (CVPR)}, 
  title={{ImageNet}: A large-scale hierarchical image database}, 
pages={248--255},
  year={2009},}

@INPROCEEDINGS{flickr30k,
  author={Plummer, Bryan A. and Wang, Liwei and Cervantes, Chris M. and Caicedo, Juan C. and Hockenmaier, Julia and Lazebnik, Svetlana},
  booktitle={Proceedings of the IEEE International Conference on Computer Vision (ICCV)}, 
  title={Flickr30k Entities: Collecting Region-to-Phrase Correspondences for Richer Image-to-Sentence Models}, 
  pages={2641--2649},
  year={2015},}

@InProceedings{coco,
author={Lin, Tsung-Yi
and Maire, Michael
and Belongie, Serge
and Hays, James
and Perona, Pietro
and Ramanan, Deva
and Doll{\'a}r, Piotr
and Zitnick, C. Lawrence},
title={Microsoft {COCO}: Common Objects in Context},
booktitle={Proceedings of the 13th European Conference on Computer Vision (ECCV)},
pages={740--755},
year={2014},
}

@article{vqa2,
  title={Mra-net: Improving vqa via multi-modal relation attention network},
  author={Peng, Liang and Yang, Yang and Wang, Zheng and Huang, Zi and Shen, Heng Tao},
  journal={IEEE Transactions on Pattern Analysis and Machine Intelligence (TPAMI)},
  pages={318--329},
  year={2020},
}

@article{itr1,
  title={Adaptive semi-supervised feature selection for cross-modal retrieval},
  author={Yu, En and Sun, Jiande and Li, Jing and Chang, Xiaojun and Han, Xian-Hua and Hauptmann, Alexander G},
  journal={IEEE Transactions on Multimedia (TMM)},
  pages={1276--1288},
  year={2018},
}

@article{itr2,
  title={Upgrading the newsroom: An automated image selection system for news articles},
  author={Liu, Fangyu and Lebret, R{\'e}mi and Orel, Didier and Sordet, Philippe and Aberer, Karl},
  journal={ACM Transactions on Multimedia Computing, Communications, and Applications (TOMM)},
  pages={1--28},
  year={2020},
}

@article{itr3,
  title={Less is better: Exponential loss for cross-modal matching},
  author={Wei, Jiwei and Yang, Yang and Xu, Xing and Song, Jingkuan and Wang, Guoqing and Shen, Heng Tao},
  journal={IEEE Transactions on Circuits and Systems for Video Technology (TCSVT)},
  year={2023},
  pages={5271-5280},
}

@INPROCEEDINGS{rq1,
  author={Chen, Zhuangzhuang and Zhang, Jin and Lai, Zhuonan and Chen, Jie and Liu, Zun and Li, Jianqiang},
    booktitle = {Proceedings of the IEEE/CVF Conference on Computer Vision and Pattern Recognition (CVPR)},
  title={Geometry-Aware Guided Loss for Deep Crack Recognition}, 
  year={2022},
  pages={4693-4702},}

@ARTICLE{lwr3,
  author={Li, Wenrui and Wang, Penghong and Xiong, Ruiqin and Fan, Xiaopeng},
  journal={IEEE Transactions on Image Processing}, 
  title={Spiking Tucker Fusion Transformer for Audio-Visual Zero-Shot Learning}, 
  year={2024},
  pages={1-1},}

@misc{zheng1,
      title={A Unified Understanding of Adversarial Vulnerability Regarding Unimodal Models and Vision-Language Pre-training Models}, 
      author={Haonan Zheng and Xinyang Deng and Wen Jiang and Wenrui Li},
      year={2024},
      eprint={2407.17797},
      archivePrefix={arXiv},
      primaryClass={cs.CV},
}

% \bibliographystyle{unsrt}

% \newpage
% \appendix

% \section{Combining IRA and TRA} \label{sec:TIRA}

% \section{Additional experiments based on CLIP} \label{sec:clip}

% \section{Generating Global Universal Adversarial Perturbation} \label{sec:global}

% \subsection{Noise Constraints}

% \subsection{Global Universal Attack Algorithms}

% \subsection{Visualization}

% \section{Additional experiments based on BEiT3} \label{sec:beit3}

% \subsection{Generating Universal Adversarial Patch}

% \subsection{Generating Universal Adversarial Global Perturbation}

\end{document}